# GI-Bench: A Panoramic Benchmark Revealing the Knowledge-Experience Dissociation of Multimodal Large Language Models in Gastrointestinal Endoscopy Against Clinical Standards

Yan Zhu[1,2,3,4]*, Te Luo[1,2,3]*, Pei-Yao Fu[1,4]*, Zhen Zhang[1,4]*, Zi-Long Wang[5], Yi-Fan Qu[1,4], Zi-Han Geng[1,4], Jia-Qi Xu[1,4], Lu Yao[1,4], Li-Yun Ma, Wei Su[1,4], Wei-Feng Chen[1,4], Quan-Lin Li[1,4], Shuo Wang[1,2,3,6], Ping-Hong Zhou[1,4]

1 Endoscopy Center and Endoscopy Research Institute, Zhongshan Hospital, Fudan University, Shanghai, 200032, China.

2 Digital Medical Research Center, School of Basic Medical Sciences, Fudan University, Shanghai, 200032, China.

3 Shanghai Key Laboratory of Medical Imaging Computing and Computer Assisted Intervention, Shanghai, 200032, China.

4 Shanghai Collaborative Innovation Center of Endoscopy, Shanghai, 200032, China.

5 Microsoft Research Asia, Shanghai, 200232, China.

6 Data Science Institute, Imperial College London, London, SW7 2AZ, UK.

## Abstract

### Background

Multimodal Large Language Models (MLLMs) show promise in gastroenterology, yet their performance against comprehensive clinical workflows and human benchmarks remains unverified.

### Objective

To systematically evaluate state-of-the-art MLLMs across a panoramic gastrointestinal endoscopy workflow and determine their clinical utility compared with human endoscopists.

### Design

We constructed GI-Bench, a benchmark encompassing 20 fine-grained lesion categories. Twelve MLLMs were evaluated across a five-stage clinical workflow: anatomical localization, lesion identification, diagnosis, findings description, and management. Model performance was benchmarked against three junior endoscopists and three residency trainees using Macro-F1, mean Intersection-over-Union (mIoU), and multi-dimensional Likert scale.

### Results

Gemini-3-Pro achieved state-of-the-art performance. In diagnostic reasoning, top-tier models

(Macro-F1 0.641) outperformed trainees (0.492) and rivaled junior endoscopists (0.727; p>0.05). However, a critical "spatial grounding bottleneck" persisted; human lesion localization (mIoU >0.506) significantly outperformed the best model (0.345; p<0.05). Furthermore, qualitative analysis revealed a "fluency-accuracy paradox": models generated reports with superior linguistic readability compared with humans (p<0.05) but exhibited significantly lower factual correctness (p<0.05) due to "over-interpretation" and hallucination of visual features. GI-Bench maintains a dynamic leaderboard that tracks the evolving performance of MLLMs in clinical endoscopy. The current rankings and benchmark results are available at https://roterdl.github.io/GIBench/.

## Conclusion

Current MLLMs function as "Advanced Beginners," possessing vast retrieved knowledge but lacking the spatial precision and clinical intuition required for independent practice. While promising as clinical copilots for drafting reports, their deployment as autonomous diagnostic agents is currently limited by significant deficits in spatial grounding and factual accuracy.

*Co-first authors

**Correspondence and reprints**

Quan-Lin Li, MD. Endoscopy Center and Endoscopy Research Institute, Zhongshan Hospital, Fudan University, Shanghai, China. E-mail: li.quanlin@zs-hospital.sh.cn

Shuo Wang, PhD. Digital Medical Research Center, School of Basic Medical Sciences, Fudan University, Shanghai, China. E-mail: shuowang@fudan.edu.cn

Ping-Hong Zhou, MD, FASGE. Endoscopy Center and Endoscopy Research Institute, Zhongshan Hospital, Fudan University, Shanghai, China. E-mail: zhou.pinghong@zs-hospital.sh.cn

## Acknowledgments

We thank Zhuo-Qi Wang, Chang Xiao and Jia-Yan Wang for their kindness and support of this research.

## Conflicts of interest

The authors have no financial, professional, or personal conflicts of interest.

## Financial support

This study was supported by the National Natural Science Foundation of China (82570629, 82270569).

**Summary**

**What is already known about this subject?**

Multimodal Large Language Models (MLLMs) have shown promise in processing medical imagery, but their application in gastroenterology has been hindered by a lack of rigorous evaluation standards. Existing benchmarks are often limited to narrow tasks (e.g., image classification) or specific organ systems, failing to reflect the comprehensive, end-to-end workflow of real-world endoscopic practice. The actual clinical utility of these models compared to human practitioners remains unverified, particularly regarding complex reasoning and decision-making capabilities.

**What are the new findings?**

We established GI-Bench, the first panoramic benchmark covering the full endoscopic workflow, revealing that top-tier MLLMs (e.g., Gemini-3-Pro) function as "Advanced Beginners," matching junior endoscopists in diagnostic reasoning but failing significantly in precise lesion localization (a "Spatial Grounding Bottleneck"). A critical "Fluency-Accuracy Paradox" was identified: MLLMs generate clinical reports with superior linguistic readability compared to human experts but exhibit significantly lower factual correctness due to hallucinations and over-interpretation of visual features. General-purpose commercial models unexpectedly outperformed domain-specific medical models in complex reasoning tasks, suggesting that current medical fine-tuning strategies primarily adapt linguistic style rather than enhancing fundamental visual-semantic understanding.

**How might it impact on clinical practice in the foreseeable future?**

Given the persistent deficits in spatial grounding and factual accuracy, current MLLMs are not ready for autonomous diagnosis but are best deployed as "clinical copilots" for drafting reports, utilizing a "Human-Directed, AI-Refined" workflow to improve efficiency while maintaining safety. This study highlights that future AI development must pivot from simple instruction tuning to addressing the "knowledge-experience dissociation," specifically improving spatial precision to enable applications in biopsy guidance and interventional endoscopy.

# Background

Gastrointestinal endoscopy is a critical tool for the diagnosis and treatment of digestive tract diseases, enabling direct visualization of the digestive tract and early detection of conditions such as colorectal cancer[1]. However, the analysis of endoscopic images heavily relies on the expertise of endoscopists, making this a high-value and high-demand clinical scenario where any improvement in efficiency and quality through artificial intelligence (AI) methods carries significant practical implications[2]. The rising global burden of gastrointestinal diseases underscores the need for advanced technologies to support accurate and timely interventions[3]. Recent advances in Multimodal Large Language Models (MLLMs), which integrate visual and linguistic modalities, offer a promising pathway for automating endoscopic analysis[4]. These models enable interactive tasks such as image captioning, visual question answering (VQA), and lesion detection, thereby potentially assisting clinicians in real-time decision-making[5]. For instance, specialized MLLMs have been developed to facilitate clinical dialogues in colonoscopy, while general-purpose models like GPT-4o demonstrate strong multimodal reasoning capabilities[6]. Despite this potential, existing benchmarks for evaluating MLLMs in endoscopy are limited in scope and clinical relevance. Current benchmarks can be broadly categorized into general-purpose medical evaluations (e.g., BLIP[7] and Qwen2-VL and domain-specific benchmarks (e.g., EndoBench[4], Kvasir-VQA[8], Surgical-VQA[9], and Gut-VLM[10]). However, these benchmarks often focus on narrow scenarios or tasks, failing to capture the full spectrum of clinical workflows.

These limitations result in benchmarks that are fundamentally decoupled from real-world clinical needs, as they fail to reflect the progressive workflow of endoscopic examination—from anatomical recognition to lesion analysis and therapeutic recommendations. Moreover, existing evaluations often prioritize image-level recognition over nuanced clinical reasoning, resulting in an incomplete assessment of MLLM capabilities.

To address these gaps, we introduce Gastrointestinal Tract-Bench (GI-Bench), a comprehensive, open-source benchmark tailored to digestive endoscopy that encompasses both gastroscopy and colonoscopy while aligning with end-to-end clinical workflows. This benchmark covers anatomical localization, lesion identification, endoscopic diagnosis, concise reporting of findings, and actionable management recommendations (Figure 1). GI-Bench is designed to answer three key questions central to the field: (1) To what extent can current MLLMs perform digestive endoscopy tasks, and which models are the most effective? (2) How do these models compare to human experts, and where do critical gaps persist? (3) How should future model development and clinical application design evolve to meet clinical needs? The principal contributions of this work include: (1) establishing the first comprehensive evaluation framework for digestive endoscopy to promote reproducibility; (2) conducting extensive experiments involving 12 MLLMs to identify model shortcomings in real-world clinical tasks; and (3) providing objective, clinically grounded performance assessments using human expert performance as a reference standard.

## Methods

**GI-Bench Dataset Construction**

To systematically evaluate state-of-the-art MLLMs on clinically meaningful gastrointestinal endoscopy tasks, we constructed GI-Bench, a benchmark dataset emphasizing clinical authenticity, expert-verified annotations, and a reproducible evaluation pipeline. Dataset development followed standard medical imaging data governance, including de-identification, expert-in-the-loop quality control, and standardized data organization to support classification and detection tasks.

This study has been reviewed and approved by the Medical Ethics Committee of Zhongshan Hospital (Approval No. b2025-145(2)). This study followed the Standards for Reporting of Diagnostic Accuracy Studies - Artificial Intelligence (STARD-AI) guidelines. All images were retrospectively collected from the existing gastrointestinal imaging archive (gastroscopy and colonoscopy) at Zhongshan Hospital. Prior to annotation, all data was de-identified, and only the minimum necessary task-related information (e.g., diagnostic information from the original reports) was retained. We curated 20 fine-grained, clinically high-impact categories across the esophagus, stomach, and colorectum: esophageal varices, reflux esophagitis, Barrett esophagus, esophageal submucosal lesion, esophageal foreign body, early esophageal cancer, advanced esophageal cancer, gastric varices, benign gastric polyp, early gastric cancer, advanced gastric cancer, gastric submucosal tumor, peptic ulcer, hyperplastic polyp, adenomatous polyp, serrated lesion, colorectal submucosal tumor, advanced colorectal cancer, colonic diverticulum, and inflammatory bowel disease. This granularity aligns with current trends toward clinically meaningful subclassification in endoscopic benchmarks and enables robust assessment of model generalization and lesion detectability.

Inclusion criteria: Static endoscopic images with sufficient quality for visual interpretation; each image contained only one of the predefined lesions. Exclusion criteria: Severe uncorrectable artifacts or occlusions (e.g., motion blur, overexposure) and any residual identifiers after de-identification. Low-quality or ambiguous samples were removed or re-annotated during quality control. All images underwent a rigorous multi-layered de-identification process to remove patient information, followed by manual verification.

We employed a two-part annotation workflow involving dual-track expert verification and consensus-based lesion localization by two experienced endoscopists (LQL and ZZ). For disease labels, each endoscopist independently assigned a single label per image, consulting contemporaneous pathology or clinical reports when available to reduce ambiguity; disagreements were resolved through consensus adjudication to yield the final label set. For lesion localization, all images containing lesions were annotated at the instance level using the open-source software LabelMe, with both experts drawing bounding boxes in a joint effort that encompassed the visible lesion boundaries while minimizing the inclusion of normal mucosa. A side-by-side review then reconciled differences in box positions and sizes to produce the final consensus bounding boxes.

To ensure a comprehensive evaluation while facilitating a feasible head-to-head comparison with human experts, we structured our analysis using two distinct dataset configurations derived from GI-Bench. First, the Full Dataset was utilized to benchmark the overall capabilities of MLLMs across all designated tasks, ensuring maximum statistical power. Second, a representative Human-AI Comparative Subset was employed, constructed by stratified random sampling from the full dataset with three images per class (60 in total). This subset was meticulously balanced to maintain the same distribution of pathological categories (e.g., polyps, ulcers, tumors) as the parent dataset, allowing for a time-efficient yet statistically robust assessment of the 'Human-AI Gap' without imposing an excessive workload on the participating endoscopists.

**Question Design**

We designed a five-question set per image to replicate the endoscopist's clinical workflow following lesion discovery: anatomical localization, lesion localization, endoscopic diagnosis, concise reporting of findings, and actionable management recommendations. The question set was crafted for automated, reproducible evaluation while constraining responses to visual evidence, thereby mitigating shortcut learning through text-only cues. Questions were originally composed in Chinese and collaboratively jointly by endoscopists to ensure clarity and alignment with local clinical practice.

The question types include multiple-choice questions (MCQ), image localization questions, and short-answer questions: Q1 (anatomical localization, MCQ) asks which gastrointestinal segment is primarily depicted; Q2 (lesion localization, structured output) requires the lesion bounding box in strict [x1, y1, x2, y2] format or "None" if absent, with no additional text; Q3 (most likely diagnosis, MCQ) requires returning only the option letter from a closed set aligned with lesion categories; Q4 ("Findings," free text) requests a 1–2 sentence description of key endoscopic features (location, morphology, color/surface, margins, and other findings relevant to disease characterization); and Q5 (follow-up recommendations, free text) seeks a 1–2 sentence, actionable next-step plan (e.g., biopsy/resection indications, surveillance intervals, or additional tests).

Gold-standard answers were established as follows: for MCQs (Q1, Q3), the disease labels established during dataset curation served as the answers; for the localization task (Q2), the standards were the expert-consensus diagnostic bounding boxes for each image; for the free-text responses (Q4, Q5), two endoscopists (LQL, ZZ) provided reference answers and consensus anchors to guide consistent scoring.

**Experimental Setup**

This study included three categories of 12 MLLMs: proprietary models (GPT-5, GPT-4o, Gemini 3 Pro, Gemini 2.5 Pro, and Claude Sonnet 4.5), open-source models (Qwen3-VL-plus,

Qwen2.5-VL-72B-Instruct, GLM-4.5V, and ERNIE-4.5-Turbo-VL), and medical open-source models (MedGemma-27b-it, HuaTuoGPT-Vision-34B, and Lingshu-32B). Detailed specifications for these models are provided in the Supplementary Materials.

Within GI-Bench, we designed and implemented differentiated prompting strategies to maximize model performance. For highly structured MCQs (Q1, Q3) and lesion localization (Q2), we strictly adopted zero-shot prompting. This setting removes inductive biases from external examples and enables a rigorous assessment of models' native capabilities—acquired during pretraining—to interpret endoscopic images, conduct logical reasoning, and select answers without human intervention. For unstructured free-text generation (Q4, Q5), given the stringent requirements for terminology accuracy, syntactic regularity, and completeness in medical reporting, we adopted few-shot prompting. Specifically, we inserted a single high-quality one-shot exemplar into the context (e.g., for Q4: "An approximately 0.8 × 0.8 cm Is-type polyp in the rectum, with a smooth surface and regular glandular architecture."). Through in-context learning, this exemplar directs models toward clinically salient features (e.g., lesion size, morphology and surface characteristics) and enforces medical formatting conventions to produce clinically useful descriptions. All models used a fixed instruction template to eliminate biases introduced by prompt engineering.

To capture each model's best-case performance while mirroring real-world cloud-based deployments, we adopted a unified API-based inference protocol. Models were evaluated using a unified inference protocol. Proprietary models were accessed via official APIs, while open-source models were deployed locally using standardized frameworks to ensure reproducibility. Comprehensive details regarding the framework architecture, specific prompt templates, and parsing algorithms are provided in the **Supplementary Materials**.

**Answer Comparison and Evaluation Methods**

Given the heterogeneous task types, we devised a stratified evaluation strategy to quantify clinical reasoning and decision-making across multiple dimensions.

For the evaluation of anatomy recognition and diagnostic classification (Q1, Q3), we employed a deterministic matching protocol that maps parsed characters to label IDs for direct comparison against clinician-defined ground truth, utilizing accuracy and F1-score as core metrics while systematically archiving raw outputs and intermediate logs to ensure transparency and facilitate downstream error analysis.

For lesion Identification (Q2), we assessed geometric consistency against expert-reviewed ground-truth bounding boxes by calculating the mean Intersection over Union (mIoU). We penalized recall failures by assigning an IoU of zero to false negatives and applied a "best-match" rule to select the highest-scoring candidate in multi-box scenarios, while simultaneously generating overlay visualizations to facilitate qualitative case reviews.

For free-text evaluation (Q4, Q5), we implemented a dual-evaluation framework that combined blinded expert review with automated LLM-as-a-judge scoring. For expert review, a panel of three experienced gastrointestinal endoscopists conducted a double-blind assessment with masked model identities and randomized case orders, rating outputs on a 5-point Likert scale across five dimensions—linguistic quality and readability, visual evidence grounding and feature coverage, factual accuracy and clinical correctness, actionability and guideline alignment, and safety and risk management—according to criteria detailed in the Supplementary Materials. Simultaneously, we employed a high-performance LLM (e.g., GPT-5) as an impartial adjudicator to produce quantitative scores by comparing model outputs against expert-written reference texts using structured prompts across these same dimensions. The reliability of this automated adjudication was rigorously validated against human expert consensus (ICC > 0.91), as detailed in Supplementary Material.

**Human Performance Evaluation**

Participants included three junior endoscopists with limited endoscopic reporting experience (fewer than 5 years of independent practice) and three residency trainees with no prior experience in endoscopic report writing. All participants underwent standardized training on task instructions and output formats prior to testing. All images were de-identified and annotated with dual-expert consensus labels and diagnostic bounding boxes. Images were presented in randomized order, with the sequence randomized independently for each participant. Each image was paired with the five-question suite described above. The interface displayed one image and its corresponding five questions at a time; no gold standards or peer answers were visible. The use of external resources was prohibited; in cases of uncertainty arising from image quality or diagnostic ambiguity, participants could express uncertainty in free-text answers, while MCQs required a single choice. Upon completion, submissions were locked to ensure auditability.

Statistical analyses were performed using custom scripts implemented in Python (Version 3.12). Descriptive statistics were used to summarize performance metrics (including Macro-Accuracy, Macro-F1, mIoU, and Likert scores) for each model across the 20 defined lesion types [doc_1]. To compare the performance of VLMs against human benchmarks (Trainee Average and Junior Average), we employed a paired sign-flip permutation test. This non-parametric approach was selected to ensure robustness given the sample size (n=20 lesion categories) and to avoid assumptions regarding data normality. The unit of analysis was the aggregated score per lesion type. A two-sided P value of less than 0.05 was considered statistically significant [doc_1]. In the results, significant differences are indicated as follows: an asterisk (*) denotes a significant difference compared with the Trainee Average, and a dagger (†) denotes a significant difference compared with the Junior Endoscopist Average.

**Data and Code Availability**

To promote transparency and facilitate future research in the field of MLLMs for gastrointestinal endoscopy, the GI-Bench dataset and its complete supporting ecosystem will be made publicly available upon publication. This ecosystem comprises the automated MLLMs evaluation framework, the human endoscopist benchmarking system, the multi-dimensional expert Likert evaluation interface, and the dynamic model leaderboard. The source code will be hosted on GitHub, and the dataset will be accessible via Google Drive.

## Results

**Performance Landscape of MLLMs in the Full Dataset**

Among all evaluated models, the commercial general-purpose models consistently achieved the highest scores across most tasks (Table 2). Gemini-3-Pro emerged as the top-performing model, establishing state-of-the-art (SOTA) performance. In anatomical recognition (Q1), it achieved a Macro-Accuracy of 0.778 and a Macro-F1 of 0.350. In the more complex disease diagnosis task (Q3), it maintained its lead with an Accuracy of 0.603 and a notable Macro-F1 of 0.601, demonstrating a balanced capability in handling both frequent and rare categories (Figure 2a-d).

Notably, in the challenging task of spatial localization (Q2), Gemini-2.5-Pro achieved the highest mIoU of 0.345, significantly outperforming other models. In contrast, GPT-4o, despite its strong reputation in general domains, exhibited unexpectedly lower performance in our specific endoscopic tasks. Its low Macro-F1 scores in both anatomical recognition (0.137) and diagnosis (0.352) suggest that its moderate accuracy is likely driven by correctly classifying majority classes, while it struggles significantly with class-imbalanced clinical scenarios. To further scrutinize the models' generalization capabilities and pinpoint specific weaknesses in fine-grained lesion analysis, the heatmap of IoU across distinct disease categories is presented in Figure 2e (see Supplementary Material for comprehensive heatmaps of all tasks).

Open-source models, such as Qwen3-VL-plus, demonstrated competitive potential, securing the second-best performance in anatomical recognition (Accuracy: 0.603; Macro-F1: 0.203) and showing robustness in diagnosis (Accuracy: 0.325; Macro-F1: 0.342). However, a critical divergence was observed in models specifically fine-tuned for the medical domain. HuatuoGPT-Vision-34B achieved a remarkable accuracy of 0.716 for anatomical recognition, rivaling top-tier commercial models. Yet, the significant discrepancy between its accuracy and its much lower Macro-F1 of 0.234 suggests that the model may be overfitting to dominant anatomical categories while failing to generalize to less common views. Furthermore, this "recognition capability" did not translate into complex clinical reasoning, as evidenced by its significantly lower diagnostic performance (Accuracy: 0.295; Macro-F1: 0.308) compared to leading generalist models. This discrepancy indicates that current medical-specific vision-language alignment may rely heavily on global image features (e.g., distinguishing a stomach form a colon) rather than the nuanced, local feature extraction required to differentiate specific diagnoses such as early gastric cancer or polyps.

To ensure the reliability of our large-scale automated assessment, we validated the GPT-5-based "impartial adjudicator" against human expert consensus using a stratified sample of model outputs. The inter-rater reliability analysis revealed an exceptional degree of alignment between the AI adjudicator and human specialists across our 5-point Likert scale metrics. Specifically, for the descriptive "Findings" (Q4) and the clinical "Subsequent Recommendations" (Q5), the Intraclass Correlation Coefficients (ICC) reached 0.912 (95% CI [0.900–0.920]) and 0.913 (95% CI [0.890–0.930]), respectively, with Pearson correlation coefficients exceeding 0.92 for both tasks (Table 4). This consistency remained robust across all five evaluation dimensions—most notably in "Safety and Risk Control" (ICC > 0.91) and "Factual Accuracy" (ICC ≈ 0.90)—statistically confirming that our GPT-5 evaluation pipeline serves as a highly accurate and scalable proxy for human expert assessment in gastrointestinal endoscopy.

For the complex reasoning and generation tasks (Q4 and Q5), a distinct performance hierarchy emerged among the evaluated MLLMs. Proprietary models demonstrated a significant advantage over open-source alternatives. Gemini-3-Pro achieved the highest performance, securing the top mean total scores for both Q4 (14.87 ± 4.83) and Q5 (13.32 ± 4.05), followed closely by Gemini-2.5-Pro. The GPT series (GPT-4o and GPT-5) formed a competitive second tier, with GPT-5 achieving a total score of 12.44 on Q4. In contrast, open-source and medically fine-tuned models, such as HuatuoGPT-Vision-34B and MedGemma-27b-it, exhibited limited capabilities in these advanced tasks, with total scores generally plateauing between 10 and 11.

A granular analysis of the evaluation dimensions reveals the specific sources of this performance gap. While most models demonstrated high proficiency in Linguistic Expression (Dimension 1), with scores consistently exceeding 3.9/5.0, significant disparities were observed in clinical grounding. The top-performing Gemini models maintained a substantial lead in Image Evidence Utilization (Dimension 2) and Safety & Risk Control (Dimension 5). For instance, in Q4, Gemini-3-Pro scored 3.15 in Dimension 2, whereas HuatuoGPT-Vision-34B scored only 1.16, indicating that while current open-source models can generate fluent text, they struggle to effectively anchor their responses in visual endoscopic findings.

**Benchmarking AI against Endoscopists in Human-AI Comparative Subset**

In the visual identification task (Q1), a substantial performance gap persisted; even the best-performing medical-specific model, HuatuoGPT-Vision-34B (Macro-F1 0.213), and the leading commercial model, GPT-5 (Macro-F1 0.187), scored significantly lower than the Junior Endoscopist Average (0.251, $P < 0.05$), highlighting the models' ongoing struggle with precise fine-grained visual feature extraction compared to human experts (Table 3, Figure 3a). Conversely, in the multiple-choice diagnostic task (Q3), top-tier MLLMs demonstrated impressive clinical reasoning capabilities that rivaled human performance. Specifically, both Gemini-3-pro and GPT-5 achieved a Macro-F1 score of 0.641, numerically surpassing the Residency Trainee Average (0.492) and showing no statistically significant difference compared to the Junior Endoscopist Average (0.727, $P > 0.05$), whereas other models like Gemini-2.5-pro (0.544) remained significantly inferior to junior experts ($P < 0.05$) (Figure 3b-d).

In the spatial grounding task (Q2), a pervasive performance gap remains between artificial intelligence and human endoscopists regarding precise lesion localization. Although gemini-2.5-pro and the medical-specific HuatuoGPT-Vision-34B achieved the highest model performance with mIoU scores of 0.385 and 0.382 respectively, they—along with all other evaluated models—exhibited statistically significant inferiority compared to both the Residency Trainee Average (0.506) and the Junior Endoscopist Average (0.543) (all $P < 0.05$) (Figure 3e). This consistent inability to match even trainee-level localization accuracy underscores a critical deficiency in current MLLMs regarding the precise coordinate-level delineation of gastrointestinal lesions, despite their relative success in semantic recognition tasks.

For the generative tasks regarding endoscopic description (Q4) and management suggestions (Q5), top-tier commercial generalist models demonstrated clinical reasoning capabilities comparable to, or even surpassing, those of human endoscopists, whereas open-source and medical-specific models still exhibited significant gaps. In Q4, Gemini-3-Pro (2.557 ± 0.750) and GPT-5 (2.417 ± 0.670) achieved performance levels with no statistical difference compared to human practitioners; conversely, medical-specific models, including HuatuoGPT-Vision-34B (2.067 ± 0.529), and the majority of open-source models scored significantly lower than both residency trainees (2.371 ± 0.368) and junior endoscopists (2.442 ± 0.504) ($P < 0.05$) (Figure 3h and Supplementary Material). Notably, in the Q5 task, Gemini-3-Pro (2.867 ± 1.039) and Gemini-2.5-Pro (2.643 ± 1.018) attained scores significantly superior to those of residency trainees (2.126 ± 0.682) and junior endoscopists (1.950 ± 0.263) ($P < 0.05$), highlighting the potential of SOTA models in assisting clinical decision-making(Figure 3i).

In the multidimensional human-model comparison (Table 4), a distinct performance dichotomy emerged between descriptive and clinical reasoning tasks. For Endoscopic

Findings (Q4), top-tier models demonstrated superior Language Expression, with GPT-4o achieving a score of 4.283 ± 0.640, significantly surpassing both Residency Trainees (3.539 ± 0.818, $P < 0.05$) and Junior Endoscopists (3.350 ± 1.040, $P < 0.05$); however, models consistently underperformed in Factual Accuracy, where Junior Endoscopists (2.944 ± 0.259) significantly outscored all commercial models, including GPT-4o (1.383 ± 0.715, $P < 0.05$) (Figure 3f). Conversely, in the Subsequent Recommendations task (Q5), Gemini-3-Pro exhibited a comprehensive advantage, achieving a Total Score of 2.867 ± 1.039 that significantly exceeded both Residency Trainees (2.126 ± 0.682, $P < 0.05$) and Junior Endoscopists (1.950 ± 0.263, $P < 0.05$) (Figure 3g). This superiority was primarily driven by higher scores in Safety (Gemini-3-Pro: 2.733 ± 1.483 vs. Trainee Average: 1.783 ± 0.660, $P < 0.05$) and Actionability, suggesting that while current VLMs may struggle with fine-grained lesion grounding, they possess robust capabilities in retrieving and applying standard clinical management guidelines.

## Discussion

The integration of Artificial Intelligence into clinical workflows has long been hindered by the lack of rigorous, workflow-aligned evaluation standards. By establishing GI-Bench, the first panoramic benchmark designed to mirror the end-to-end cognitive process of digestive endoscopy—we provide a definitive assessment of the current capabilities of MLLMs. Our comprehensive evaluation reveals that while state-of-the-art models have achieved a pivotal milestone, they currently occupy the level of "Advanced Beginners" within the clinical domain[11]. Top-tier models, particularly Gemini-3-Pro, demonstrated anatomical recognition capabilities that rival those of junior endoscopists, and diagnostic reasoning that significantly outperforms that of residency trainees. However, a critical performance chasm remains when these models are pitted against independent junior endoscopists, particularly in tasks requiring nuanced spatial grounding and precise lesion characterization. This disparity suggests that current MLLMs exhibit a "knowledge-experience dissociation"[12]: much like a medical student who has mastered textbook theory but lacks the tacit knowledge derived from clinical practice, these models possess vast encyclopedic retrieval abilities yet falter in the intuitive pattern recognition required for independent decision-making[13]. Consequently, while MLLMs show immense promise as assistive tools, they are not yet equipped to function as autonomous diagnostic agents in complex endoscopic scenarios.

While MLLMs demonstrate emerging proficiency in semantic classification, our results expose a severe "Spatial Grounding Bottleneck" that currently precludes their deployment in interventional settings. A profound disconnect exists between the models' ability to identify what a pathology is (Diagnosis Q3) and their inability to pinpoint where it is (Localization Q2). Humans consistently achieved a mean Intersection over Union (mIoU) exceeding 0.5, whereas generalist models like GPT-4o failed catastrophically (mIoU 0.033), and even the best-performing Gemini-2.5-Pro (mIoU 0.345) lagged significantly behind trainees. This failure likely stems from the architectural limitations of current Vision Transformers, which struggle to align high-level semantic tokens with local, fine-grained pixel features in high-resolution medical imagery[14]. The inability to provide precise coordinates is not merely a technical deficit but a fundamental clinical barrier; without accurate spatial localization, AI cannot guide biopsies, assist in endoscopic resection, or flag regions of interest for robotic intervention[15]. Furthermore, this spatial blindness compromises the interpretability of generated reports, as the model cannot verify whether its textual description of a "0.8-cm polyp" actually corresponds to the visual pixels of that polyp, leading to ungrounded and potentially hazardous clinical narratives.

Perhaps the most insidious finding of our study is the "Fluency-Accuracy Paradox," where MLLMs generate clinical reports that are linguistically superior to those of human experts yet factually unreliable. Our qualitative analysis reveals that models like GPT-4o achieve near-perfect readability scores by adopting a polite, structured, and grammatically flawless tone. However, this linguistic fluency often masks "over-interpretation," where models hallucinate non-existent visual evidence—such as specific vascular patterns or mucosal textures—to create a comprehensive-sounding narrative[16]. In a clinical context, this "convincing but wrong" output is far more dangerous than an obvious error, as it mimics the authoritative style of a senior consultant, potentially bypassing the cognitive vigilance of junior endoscopists. Additionally, we observed a prevalence of "defensive medicine" in AI recommendations (Q5), where models indiscriminately listed every possible therapeutic option to maximize safety scores, lacking the precision to tailor decisions to the specific case. Conversely, human brevity often results from reliance on mental templates and implicit knowledge, which models currently lack. These findings suggest that the immediate future of AI in endoscopy lies not in full automation, but in a "Human-Directed, AI-Refined" collaborative workflow. In this paradigm, the endoscopist provides the critical diagnostic direction and key findings (ensuring factual accuracy), while the MLLM functions as an advanced scribe, expanding these keywords into a fluent, standardized, and comprehensive clinical report[17].

Our findings reveal a counterintuitive landscape where generalist models (e.g., Gemini-3-Pro, GPT-4o) frequently outperform domain-specific models (e.g., HuatuoGPT) in complex diagnostic reasoning, despite the latter's explicit fine-tuning on medical corpora (see Supplementary Material for detailed cost-effectiveness analysis). This performance gap suggests that current instruction-tuning strategies for medical MLLMs may function primarily as style alignment rather than genuine medical knowledge injection. While domain-specific fine-tuning successfully adapts the model's tone to a medical persona—as evidenced by

HuatuoGPT's strong performance in anatomical recognition—it appears insufficient to instill the deep visual-semantic understanding required for fine-grained lesion characterization[12]. Generalist models, conversely, benefit from massive-scale pre-training that fosters emergent reasoning capabilities and broad world knowledge, allowing them to generalize to endoscopic tasks in a zero-shot setting with superior efficacy[18]. This observation aligns with recent assertions that the "reasoning engine" of a foundation model is established during pre-training, and that superficial fine-tuning cannot compensate for a lack of fundamental visual representations[11]. Consequently, future development should likely pivot away from simple instruction tuning towards injecting medical multimodal data during the pre-training phase or adopting Retrieval-Augmented Generation frameworks to bridge the knowledge gap without sacrificing the reasoning power of large generalist baselines[19].

Our study has several limitations that merit consideration. First, GI-Bench relies exclusively on static imagery, which inherently decouples the evaluation from the temporal dynamics of real-time endoscopy. This static approach may obscure diagnostic cues derived from motion, tissue manipulation, and multi-angle observation, as temporal information is often essential for distinguishing between artifacts and genuine lesions in multimodal biomedical AI. Second, while our dataset encompasses 20 fine-grained lesion categories, its single-center origin may introduce geographic or institutional biases regarding patient demographics and device manufacturers. Future iterations must expand to multi-center cohorts to ensure robust generalization across diverse clinical environments.

Looking forward, the evolution of endoscopic AI must transcend static analysis. Future research should prioritize the development of video-based MLLMs capable of processing temporal streams to enhance lesion tracking and surgical phase recognition. Furthermore, our findings regarding the "Fluency-Accuracy Paradox" where models exhibit superior linguistic readability but inferior factual correctness compared to humans—suggest a critical pivot in deployment strategy. Rather than functioning as autonomous diagnosticians, MLLMs are best positioned as "clinical copilots". By leveraging their superior linguistic fluency for report drafting while relying on human experts for factual verification and decision-making, this human-in-the-loop collaboration can maximize efficiency while mitigating the risks of hallucination and inaccuracy identified in our study.

In conclusion, GI-Bench serves as a rigorous reality check: while current MLLMs have mastered the "textbook" knowledge of endoscopy, they have not yet acquired the "clinical intuition" of an experienced physician, highlighting the need for a symbiotic, rather than substitutive, AI-physician partnership.

# Figure Legends

Figure 1: Overview of the GI-Bench study design and evaluation framework.

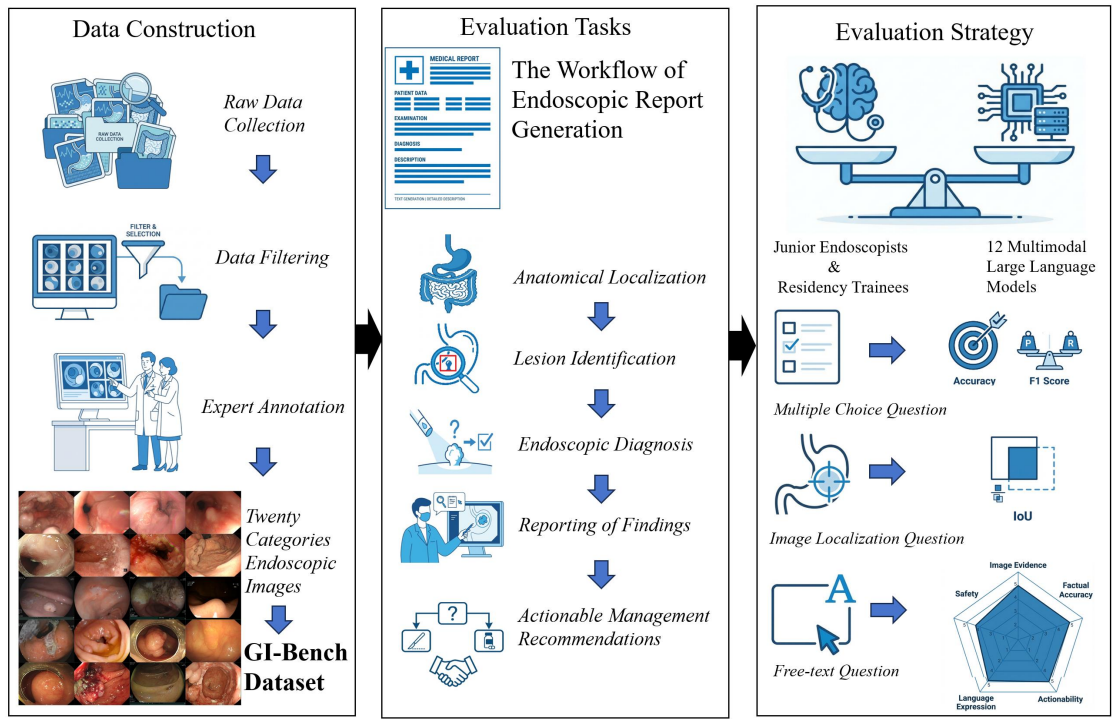

Figure 2: Performance evaluation of MLLMs across anatomical localization and fine-grained disease diagnosis.

a, Bar chart displaying the F1 scores for the anatomical localization task (Q1).

b–d, Comparative analysis of F1 scores across multi-choice diagnostic task (Q3).

e, Heatmap of IoU scores across distinct disease categories for the task of spatial localization (Q2).

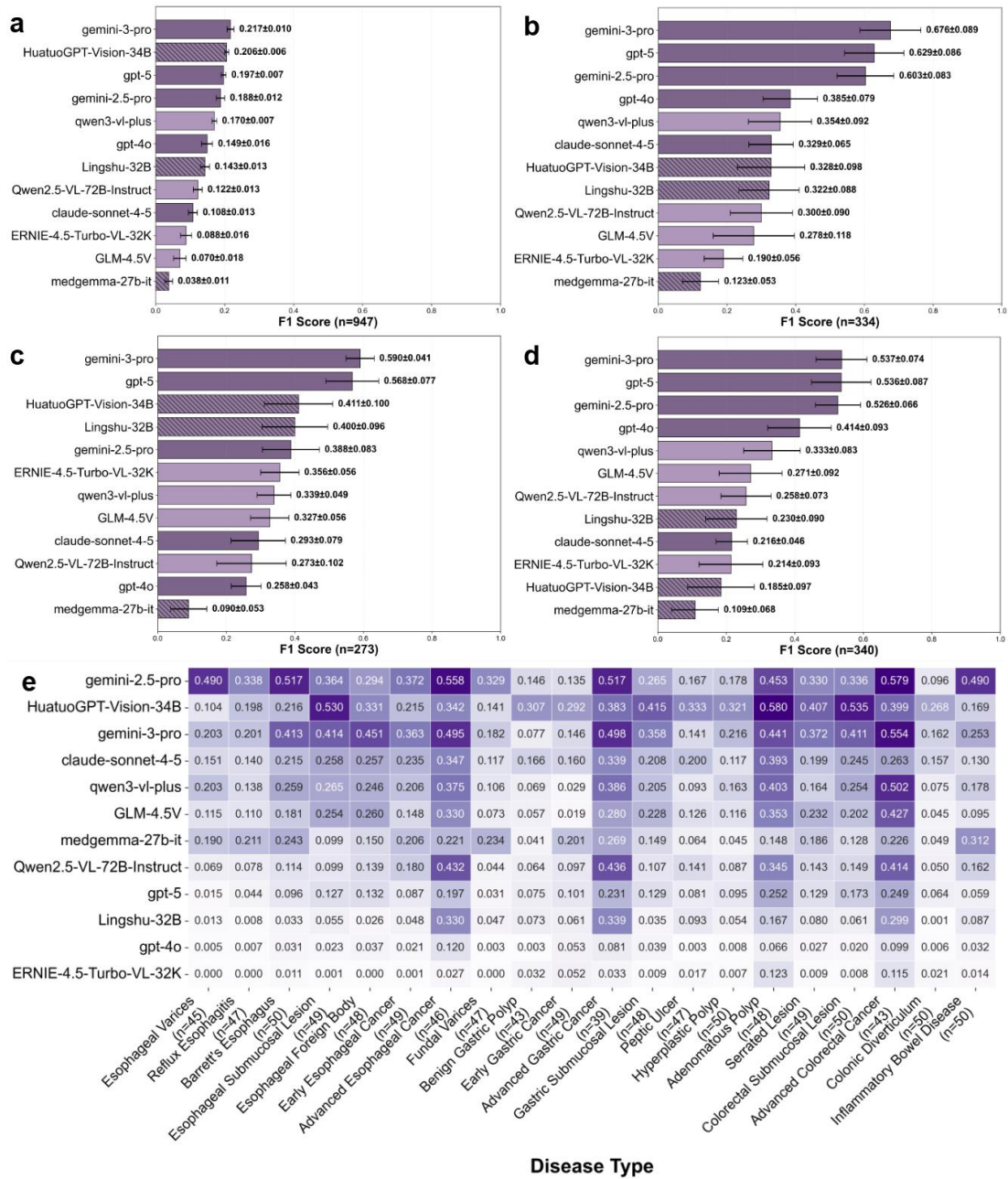

Figure 3 Benchmarking MLLMs performance against human endoscopists across multimodal clinical tasks.

a, Bar chart comparing F1 scores for the anatomical localization task (Q1).

b–d, Comparative analysis of F1 scores for diagnostic tasks (Q3) stratified by anatomical regions (esophagus, stomach, and colorectum).

e, Heatmap of mIoU for lesion localization (Q2).

f, g, Radar charts illustrating the multidimensional evaluation of free-text tasks: endoscopic findings description (Q4) and management recommendations (Q5).

h, i, Violin plots displaying the distribution of total quality scores for tasks Q4 and Q5,

respectively.

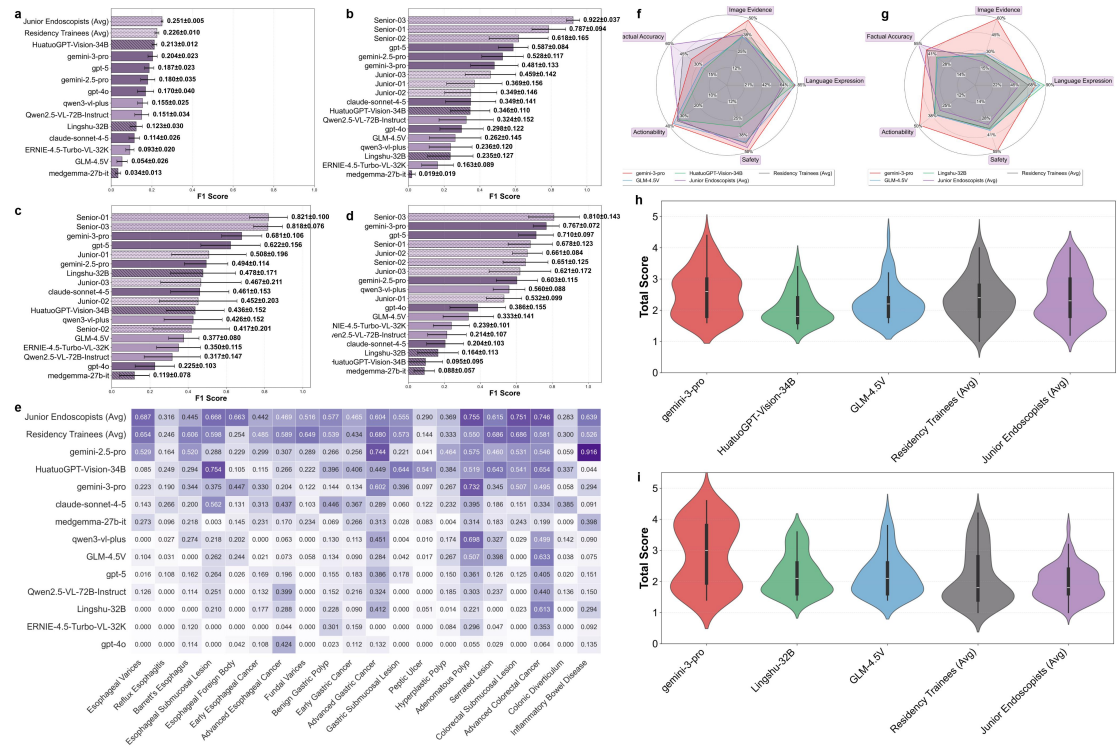

Table 1. Comparison of endoscopic lesion assessment tasks and output formats: Polished (constrained) vs. Original (open-ended)

| Task Type | Polished (Constrained Output) | Original (Open-ended) |
| --- | --- | --- |
| Q1 (anatomical localization, MCQ) | Which segment of the GI tract does the main anatomical part shown in this image belong to? Please choose from the following options and return only the letter representing the answer, no other text. Options: A. Esophagus, B. Stomach, C. Duodenum, D. Colorectum. | Which segment of the GI tract does the main anatomical part shown in this image belong to? (Choose from Esophagus, Stomach, Duodenum, or Colorectum) |
| Q2 (lesion identification, structured output) | [Instruction] Analyze the image. Task: Output the bounding box of the lesion in the image. Output Rule: 1. If no lesion, output: None. 2. If there is a lesion, output must be and only be the bounding box coordinates in the correct format: [x1, y1, x2, y2]. No other text allowed. | Please annotate the bounding box coordinates of the only lesion in the image (Format: [x1, y1, x2, y2]), and briefly explain the image evidence for the annotation. |
| Q3 (most likely diagnosis, MCQ) | What is the most likely diagnosis for this lesion? Please choose from the following options and return only the letter representing the answer, no other text. Options: A. Esophageal Varices, B. Reflux Esophagitis, C. Barrett's Esophagus, D. Esophageal Submucosal Tumor, E. Esophageal Foreign Body, F. Early Esophageal Cancer, G. Advanced Esophageal Canc | What is the most likely diagnosis for this lesion? (A. Esophageal Varices, B. Reflux Esophagitis, C. Barrett's Esophagus, D. Esophageal Submucosal Tumor, E. Esophageal Foreign Body, F. Early Esophageal Cancer, G. Advanced Esophageal Cancer … [List continues for Stomach/Colon options])? And state the basis for the inferen |

| Task Type | Polished (Constrained Output) | Original (Open-ended) |
|---|---|---|
| | er, H. None. | ce and probability. |
| Q4 ("Findings," free text) | Acting as an endoscopy expert, write the "Findings" section of the report in 1-2 sentences. Requirements: Concise description, highlighting core morphological features of the lesion (e.g., location, size, Paris classification, surface structure, etc.). Example: "A type Is polyp about 0.8*0.8cm is seen in the rectum, with a smooth surface and regular pit pattern." | Write the "Findings" section of the report. Requirements: Concise description, highlighting core morphological features of the lesion (e.g., location, size, Paris classification, surface structure, etc.). Example: "A type Is polyp about 0.8*0.8cm is seen in the rectum, with a smooth surface and regular pit pattern." |
| Q5 (follow-up recommendations, free text) | Acting as an endoscopy expert, write the "Follow-up Advice" section of the report in 1-2 sentences. Requirements: Concise description, clear instructions, including necessary next steps for diagnosis and treatment. Example: "Endoscopic en bloc resection (ESD preferred, EMR/snare resection optional based on base width and pedicle) is recommended under no contraindications; send for pathology to evaluate margins; if pathology suggests high-grade intraepithelial neoplasia/invasive cancer or positive margins, add treatment and follow-up according to guidelines. Regular colonoscopy review pending pathology results." | For this lesion in the image, please briefly explain the follow-up treatment plan and its rationale. |

Table 2. Performance of 12 commercial and open-source multimodal large language models in the full dataset on gastrointestinal endoscopic tasks: multiple choice questions, lesion localization, and free text generation.

| Trial-participating model | | Multiple Choice Question | | | | Lesion Localization | | | Free Text | |
|---|---|---|---|---|---|---|---|---|---|---|
| Multimodal Large Language Models | | Q1 | | Q3 | | Q2 | | | Q4 | Q5 |
| | | Macro-Acc | Macro-F1 | Macro-Acc | Macro-F1 | mIoU | Recall@0.5 | Recall@0.75 | Mean ± SD | Mean ± SD |
| Commercial Models | gpt-5 | 0.659 | 0.207 | 0.551 | 0.577 | 0.117 | 0.039 | 0.000 | 2.414±0.616 | 2.487±0.696 |
| | gpt-4o | 0.263 | 0.137 | 0.349 | 0.352 | 0.033 | 0.002 | 0.000 | 2.298±0.557 | 2.468±0.726 |
| | claude-sonnet-4-5 | 0.36 | 0.128 | 0.303 | 0.279 | 0.213 | 0.138 | 0.014 | 2.043±0.517 | 2.028±0.675 |
| | gemini-2.5-pro | 0.574 | 0.231 | 0.542 | 0.506 | 0.345 | 0.326 | 0.08 | 2.663±0.909 | 2.869±1.018 |
| | gemini-3-pro | 0.778 | 0.350 | 0.603 | 0.601 | 0.316 | 0.281 | 0.057 | 2.664±0.811 | 2.974±0.966 |
| Open-source | Qwen2.5-VL-72B-Instruct | 0.402 | 0.164 | 0.323 | 0.277 | 0.164 | 0.144 | 0.04 | 2.028±0.353 | 2.089±0.594 |

| Trial-participating model | | Multiple Choice Question | | | | Lesion Localization | | | Free Text | |
|---|---|---|---|---|---|---|---|---|---|---|
| | Multimodal Large Language Models | Q1 | | Q3 | | Q2 | | | Q4 | Q5 |
| Models | Qwen3-VL-plus | 0.603 | 0.203 | 0.325 | 0.342 | 0.213 | 0.184 | 0.029 | 2.151±0.535 | 2.225±0.678 |
| | GLM-4.5V | 0.333 | 0.269 | 0.319 | 0.292 | 0.181 | 0.144 | 0.02 | 2.239±0.539 | 2.283±0.672 |
| | ERNIE-4.5-Turbo-VL-32K | 0.468 | 0.202 | 0.304 | 0.253 | 0.023 | 0.016 | 0.011 | 2.081±0.466 | 2.154±0.591 |
| Medical Open-source Models | HuatuoGPT-Vision-34B | 0.716 | 0.234 | 0.295 | 0.308 | 0.324 | 0.293 | 0.062 | 2.076±0.471 | 2.029±0.554 |
| | Lingshu-32B | 0.480 | 0.236 | 0.308 | 0.317 | 0.092 | 0.062 | 0.008 | 2.020±0.466 | 2.244±0.673 |
| | medgemma-27b-it | 0.197 | 0.118 | 0.193 | 0.107 | 0.168 | 0.045 | 0.001 | 2.008±0.489 | 2.022±0.604 |

Table 3. Table 3. Performance comparison between 12 multimodal large language models and human endoscopists (junior endoscopists and residency trainees) in Human-AI comparative subset.

| Participants | | Q1 | | Q3 | | Q2 | | | Q4 | Q5 |
|---|---|---|---|---|---|---|---|---|---|---|
| | | Macro-Acc | Macro-F1 | Macro-Acc | Macro-F1 | mIoU | Recall@0.5 | Recall@0.75 | Mean ± SD | Mean ± SD |
| Commercial Models | gpt-5 | 0.617*† | 0.187† | 0.617 | 0.641 | 0.159*† | 0.067*† | 0.000*† | 2.417±0.670 | 2.367±0.718† |
| | gpt-4o | 0.300*† | 0.170 | 0.333† | 0.307*† | 0.062*† | 0.017*† | 0.000*† | 2.310±0.594 | 2.350±0.696† |
| | claude-sonnet-4-5 | 0.367*† | 0.114*† | 0.383† | 0.332† | 0.261*† | 0.200*† | 0.000*† | 2.000±0.522*† | 1.897±0.608* |
| | gemini-2.5-pro | 0.550*† | 0.180 | 0.600 | 0.544† | 0.385*† | 0.383*† | 0.133† | 2.530±0.978 | 2.643±1.018*† |
| | gemini-3-pro | 0.700† | 0.204 | 0.667 | 0.641 | 0.315*† | 0.267*† | 0.050*† | 2.557±0.750 | **2.867±1.039*†** |
| Open-source Models | Qwen2.5-VL-72B-Instruct | 0.483*† | 0.151† | 0.350† | 0.283† | 0.158*† | 0.100*† | 0.000*† | 1.963±0.367*† | 2.057±0.587 |
| | Qwen3-VL-plus | 0.533*† | 0.155*† | 0.400† | 0.407† | 0.173*† | 0.167*† | 0.033*† | 2.087±0.537*† | 2.143±0.623 |
| | GLM-4.5V | 0.317*† | 0.054*† | 0.367† | 0.321† | 0.164*† | 0.150*† | 0.033*† | 2.250±0.599 | 2.207±0.689† |
| | ERNIE-4.5-Trube-VL-32K | 0.483*† | 0.093*† | 0.283*† | 0.246*† | 0.075*† | 0.050*† | 0.050*† | 2.033±0.370*† | 2.077±0.546 |
| Medical Open-source Models | HuatuoGPT-Vision-34B | 0.767† | 0.213† | 0.283*† | 0.285† | 0.382*† | 0.367† | 0.133† | 2.067±0.529*† | 1.967±0.600 |
| | Lingshu-32B | 0.467*† | 0.123*† | 0.317† | 0.283† | 0.131 | 0.117*† | 0.017*† | 2.007±0.45 | 2.237±0.65 |

| Participants | | Q1 | | Q3 | | Q2 | | | Q4 | Q5 |
|---|---|---|---|---|---|---|---|---|---|---|
| | | Macro-Acc | Macro-F1 | Macro-Acc | Macro-F1 | mIoU | Recall@0.5 | Recall@0.75 | Mean ± SD | Mean ± SD |
| | | | | | | *† | | | 7*† | 9† |
| | medgemma-27b-it | 0.217*† | 0.034*† | 0.150*† | 0.073*† | 0.174*† | 0.017*† | 0.000*† | 1.983±0.526*† | 1.877±0.626* |
| Junior Endoscopists | Junior 1 | 0.883 | 0.235 | 0.767 | 0.759 | **0.643** | **0.750** | **0.483** | 2.680±0.733 | 2.123±0.483 |
| | Junior 2 | 0.983 | 0.256 | 0.583 | 0.569 | 0.521 | 0.500 | 0.267 | **2.783±0.735** | 2.080±0.627 |
| | Junior 3 | **1.000** | **0.261** | **0.867** | **0.852** | 0.464 | 0.517 | 0.100 | 1.863±0.342 | 1.647±0.350 |
| Residency Trainees | Trainee 1 | 0.817 | 0.215 | 0.500 | 0.489 | 0.529 | 0.650 | 0.167 | 1.970±0.563 | 1.437±0.358 |
| | Trainee 2 | 0.883 | 0.231 | 0.533 | 0.518 | 0.458 | 0.483 | 0.200 | 2.450±0.578 | 2.140±0.691 |
| | Trainee 3 | 0.883 | 0.231 | 0.500 | 0.468 | 0.530 | 0.550 | 0.317 | 2.693±0.810 | 2.800±0.892 |
| Junior Endoscopists | Junior Average | **0.956*** | 0.251 | 0.739* | 0.727* | **0.543** | 0.589 | 0.283 | 2.442±0.504 | 1.950±0.263* |
| Residency Trainees | Trainee Average | 0.861† | 0.226 | 0.511† | 0.492† | 0.506 | 0.561 | 0.228 | 2.371±0.368 | 2.126±0.682† |

\* P < 0.05$ compared with Residency Trainee Average.

† P < 0.05$ compared with Junior Endoscopist Average.

mIoU, mean Intersection over Union; SD, standard deviation.

Table 4. Comparative performance on colonoscopy QA (Q4 findings) and subsequent recommendations (Q5): per-dimension means and standard deviations (language expression, image evidence, factual accuracy, actionability, safety) and total scores for models and clinical participants.

| | | | Q4 (Findings) | | | | | |
|---|---|---|---|---|---|---|---|---|
| Participants | | n | language expression (Mean±SD) | image evidence (Mean±SD) | factual accuracy (Mean±SD) | actionability (Mean±SD) | safety (Mean±SD) | Total Score (Mean±SD) |
| Commercial Models | claude-sonnet-4-5 | 60 | 3.933±0.482*† | 1.533±0.769*† | 1.133±0.468*† | 1.600±0.694 | 1.800±0.898*† | 2.000±0.522*† |
| | gemini-2.5-pro | 60 | 3.617±0.585† | 2.117±1.091 | 1.750±1.144* | **2.400±1.153*†** | **2.767±1.533*** | 2.530±0.978 |
| | gemini-3-pro | 60 | 4.083±0.381*† | 2.433±0.890*† | 1.867±1.065† | 1.917±0.743 | 2.483±1.282 | 2.557±0.750 |
| | gpt-4o | 60 | 4.283±0.640*† | 1.950±0.832 | 1.383±0.715*† | 1.733±0.710 | 2.200±1.102 | 2.310±0.594 |

## Q4 (Findings)

| Participants | | n | language expression (Mean±SD) | image evidence (Mean±SD) | factual accuracy (Mean±SD) | actionability (Mean±SD) | safety (Mean±SD) | Total Score (Mean±SD) |
|---|---|---|---|---|---|---|---|---|
| | gpt-5 | 60 | 3.983±0.469*† | 2.067±0.954 | 1.600±0.995*† | 1.900±0.656 | 2.533±1.065 | 2.417±0.670 |
| Open-source Models | ERNIE-4.5-Turbo-VL-32K | 60 | 4.033±0.520*† | 1.650±0.709* | 1.100±0.354*† | 1.433±0.533*† | 1.950±0.769† | 2.033±0.370*† |
| | GLM-4.5V | 60 | 4.133±0.430*† | 1.800±0.840* | 1.317±0.770*† | 1.867±0.700 | 2.133±1.033 | 2.250±0.599 |
| | Qwen2.5-VL-72B-Instruct | 60 | 3.900±0.354*† | 1.567±0.673*† | 1.100±0.440*† | 1.250±0.474*† | 2.000±0.689 | 1.963±0.367*† |
| | qwen3-vl-plus | 60 | 4.250±0.628*† | 1.767±0.722* | 1.200±0.546*† | 1.533±0.623 | 1.683±1.000*† | 2.087±0.537*† |
| Medical Open-source Models | HuatuoGPT-Vision-34B | 60 | 4.017±0.701*† | 1.883±0.825 | 1.183±0.431*† | 1.683±0.701 | 1.567±0.810*† | 2.067±0.529*† |
| | Lingshu-32B | 60 | **4.467±0.566*†** | 1.650±0.820* | 1.150±0.515*† | 1.400±0.527*† | 1.367±0.688*† | 2.007±0.457*† |
| | medgemma-27b-it | 60 | 3.133±0.503* | 1.717±0.739* | 1.183±0.567*† | 1.667±0.572 | 2.217±0.940 | 1.983±0.526*† |
| Junior Endoscopists | Junior 1 | 60 | 2.150±0.444 | 1.083±0.279 | **3.233±1.015** | 1.083±0.279 | 1.767±0.465 | 1.863±0.342 |
| | Junior 2 | 60 | 3.917±0.696 | 2.383±0.691 | 2.867±1.567 | 2.050±0.699 | 2.700±0.997 | **2.783±0.735** |
| | Junior 3 | 60 | 3.983±0.770 | 2.183±0.624 | 2.733±1.313 | 1.950±0.699 | 2.550±0.964 | 2.680±0.733 |
| Residency Trainees | Trainee 1 | 60 | 3.483±0.873 | 2.167±0.587 | 2.517±1.242 | 1.783±0.555 | 2.300±0.809 | 2.450±0.578 |
| | Trainee 2 | 60 | 4.383±0.613 | **2.517±0.930** | 2.083±1.306 | 2.083±0.787 | 2.400±1.153 | 2.693±0.810 |
| | Trainee 3 | 60 | 2.750±0.751 | 1.633±0.637 | 2.283±1.223 | 1.350±0.481 | 1.833±0.763 | 1.970±0.563 |
| Junior Endoscopists | Junior Average | 180 | 3.350±1.040* | 1.883±0.700* | 2.944±0.259* | 1.694±0.532 | 2.339±0.501 | 2.442±0.504 |
| Residency Trainees | Trainee Average | 180 | 3.539±0.818† | 2.106±0.445† | 2.294±0.217† | 1.739±0.369 | 2.178±0.303 | 2.371±0.368 |

## Q5 (Subsequent Recommendations)

| Participants | | n | language expression (Mean±SD) | image evidence (Mean±SD) | factual accuracy (Mean±SD) | actionability (Mean±SD) | safety (Mean±SD) | Total Score (Mean±SD) |
|---|---|---|---|---|---|---|---|---|
| Commercial Models | claude-sonnet-4-5 | 60 | 3.717±0.524*† | 1.333±0.542† | 1.417±0.787*† | 1.483±0.833 | 1.533±0.929 | 1.897±0.608* |
| | gemini-2.5-pro | 60 | 3.383±0.585† | 2.917±1.211*† | 2.183±1.359† | 2.217±1.027*† | 2.517±1.444* | 2.643±1.018*† |
| | gemini-3-pro | 60 | 3.850±0.515*† | **2.967±1.089*†** | 2.350±1.424 | **2.433±1.240*†** | 2.733±1.483*† | **2.867±1.039*†** |
| | gpt-4o | 60 | 3.983±0.390*† | 2.150±0.840*† | 1.783±0.993*† | 1.883±0.846† | 1.950±1.080† | 2.350±0.696† |
| | gpt-5 | 60 | 3.983±0.390*† | 2.233±0.810*† | 1.733±0.954*† | 1.967±0.901† | 1.917±1.078 | 2.367±0.718† |
| Open-source Models | ERNIE-4.5-Turbo-VL-32K | 60 | 3.917±0.497*† | 1.600±0.588*† | 1.450±0.852*† | 1.717±0.739 | 1.700±0.889 | 2.077±0.546 |

| Participants | | n | language expression (Mean±SD) | image evidence (Mean±SD) | factual accuracy (Mean±SD) | actionability (Mean±SD) | safety (Mean±SD) | Total Score (Mean±SD) |
|---|---|---|---|---|---|---|---|---|
| | | | | | | | | |
| | GLM-4.5V | 60 | 4.117±0.691*† | 1.450±0.594† | 1.800±0.971*† | 1.800±0.777 | 1.867±1.016 | 2.207±0.689† |
| | Qwen2.5-VL-72B-Instruct | 60 | 4.017±0.651*† | 1.283±0.454† | 1.667±0.914*† | 1.650±0.755 | 1.667±0.914 | 2.057±0.587 |
| | qwen3-vl-plus | 60 | 4.433±0.593*† | 1.433±0.593† | 1.550±0.852*† | 1.733±0.861 | 1.567±0.890 | 2.143±0.623 |
| Medical Open-source Models | HuatuoGPT-Vision-34B | 60 | 3.767±0.647*† | 1.217±0.415 | 1.567±0.851*† | 1.617±0.739 | 1.667±0.933 | 1.967±0.600 |
| | Lingshu-32B | 60 | **4.417±0.530*†** | 1.317±0.469† | 1.867±1.081*† | 1.767±0.810 | 1.817±1.000 | 2.237±0.659† |
| | medgemma-27b-it | 60 | 2.950±0.387* | 1.900±1.145*† | 1.400±0.694*† | 1.567±0.673 | 1.567±0.810 | 1.877±0.626* |
| Junior Endoscopists | Junior 1 | 60 | 2.183±0.537 | 1.000±0.000 | 2.450±0.852 | 1.283±0.490 | 1.317±0.469 | 1.647±0.350 |
| | Junior 2 | 60 | 2.917±0.869 | 1.133±0.389 | 2.900±1.037 | 1.617±0.715 | 1.833±0.867 | 2.080±0.627 |
| | Junior 3 | 60 | 3.300±0.720 | 1.033±0.181 | **2.850±1.055** | 1.700±0.619 | 1.733±0.634 | 2.123±0.483 |
| Residency Trainees | Trainee 1 | 60 | 3.683±0.911 | 1.217±0.490 | 2.367±1.178 | 1.700±0.809 | 1.733±0.821 | 2.140±0.691 |
| | Trainee 2 | 60 | 4.600±0.558 | 1.783±0.739 | 2.767±1.332 | 2.383±1.091 | 2.467±1.214 | 2.800±0.892 |
| | Trainee 3 | 60 | 1.867±0.566 | 1.083±0.279 | 1.850±0.840 | 1.233±0.465 | 1.150±0.404 | 1.437±0.358 |
| Junior Endoscopists | Junior Average | 180 | 2.800±0.568* | 1.055±0.069* | 2.733±0.247* | 1.533±0.221 | 1.628±0.274 | 1.950±0.263* |
| Residency Trainees | Trainee Average | 180 | 3.383±1.391† | 1.361±0.372† | 2.328±0.460† | 1.772±0.578† | 1.783±0.660 | 2.126±0.682† |

## A.1 Experimental setup

### A.1.1 Model Selection

We evaluated 12 state-of-the-art multimodal large language models (MLLMs), encompassing both proprietary and open-source systems, to comprehensively benchmark the performance boundaries of current technologies.

**Proprietary Models.** GPT-5, GPT-4o, Gemini 3 Pro, Gemini 2.5 Pro, and Claude Sonnet 4.5. Developed by leading AI laboratories, these models represent the forefront of closed-source technology, leveraging massive-scale computational resources and datasets. Selection was based on their superior reasoning, perception, and instruction-following capabilities demonstrated on general visual question-answering (VQA) benchmarks (e.g., MMMU, MMBench, MathVista). Our objective was to assess their zero-shot generalization potential in specialized medical domains without domain-specific fine-tuning.

**Open-Source Models.** Qwen3-VL-plus, Qwen2.5-VL-72B-Instruct, GLM-4.5V, and ERNIE-4.5-Turbo-VL. Developed primarily by leading Chinese technology enterprises (Alibaba, Zhipu AI, and Baidu), these models represent the open-source community's latest efforts to achieve parity with top-tier proprietary systems. Notably, their training corpora incorporate substantial Chinese medical literature, domestic clinical guidelines, and imaging data featuring Asian demographic characteristics, conferring potential advantages in Chinese language comprehension, localized medical terminology interpretation, and cultural adaptability. Evaluation of these models is essential for advancing open-source research and facilitating cost-effective clinical deployment.

**Medical Open-Source Models.** MedGemma-27b-it, HuaTuoGPT-Vision-34B, and Lingshu-32B. These models exemplify the domain-specific specialization paradigm. Built upon general-purpose foundation models, they have undergone extensive instruction tuning and alignment using large-scale medical corpora, including textbooks, clinical guidelines, de-identified electronic health records (EHRs), and medical imaging datasets. Specifically, MedGemma-27b-it emphasizes multimodal fusion through targeted pre-training on medical images and EHRs to bridge the semantic gap between clinical text and imaging modalities. Lingshu-32B demonstrates superior modality coverage, supporting over 12 major medical imaging types spanning radiology, pathology, and specialty domains such as endoscopy. HuaTuoGPT-Vision has been optimized using a curated dataset comprising 1.3 million medical visual question-answer pairs. Evaluation of these models serves to validate the efficacy of domain-specific fine-tuning in mitigating hallucinations and enhancing diagnostic accuracy, while exploring the potential for medium-scale specialized models to outperform larger general-purpose models in high-stakes medical applications.

### A.1.2 Automated Evaluation Framework

To ensure efficient, standardized, and fully reproducible large-scale evaluation, we developed a modular Python-based automation framework comprising five core components.

**Multimodal Data Loader.** Standardizes heterogeneous data sources for efficient ingestion of images and textual queries, incorporating a built-in preprocessing pipeline to satisfy each model's tensor and encoding specifications.

**Adaptive Prompt Constructor.** Dynamically assembles prompts according to task type, integrating system instructions, few-shot exemplars, and the current query to ensure structured and consistent model inputs.

**Unified Model Client.** An abstraction layer encapsulating heterogeneous API protocols. To mitigate service instability, we implemented fault-tolerant mechanisms with exponential backoff retry logic (maximum of two attempts) for network timeouts and server errors (e.g., HTTP 500, 502), thereby minimizing interruptions from transient failures.

**Structured Output Parser.** A rule-based postprocessor that transforms unstructured model outputs into standardized, quantifiable formats (e.g., option labels, bounding box coordinates) using regular expressions and coordinate extraction algorithms.

**Comprehensive Logging System.** Granular logging of each API interaction, capturing input prompts, raw responses, parsed outputs, and temporal metadata to facilitate systematic error analysis and ensure reproducibility.

### A.1.3 Input Standardization and Prompt Template Design

**Image Preprocessing.** Given the substantial heterogeneity in fields of view across endoscope manufacturers and models, raw images frequently contain non-diagnostic black borders. To address this, we implemented an automated region-of-interest (ROI) extraction algorithm that cropped valid fields of view while discarding extraneous background. All preprocessed images were subsequently serialized to Base64 format as standardized visual inputs for the VLMs, thereby ensuring consistent input distributions.

**Prompt Template Design.** To ensure clinical professionalism, formatting regularity, and output parsability, we developed a structured prompting framework comprising a unified system prompt and task-specific output constraints.

**System Prompt (Persona Setting).** At the initiation of each interaction, all models received the following instruction: "You are a professional gastrointestinal endoscopist capable of accurately analyzing endoscopic images and providing expert insights." This prompt established the clinical imaging context and activated domain-relevant internal knowledge representations.

**Task-Specific Constraints:**

- Multiple-choice questions (Q1, Q3). A single-character output constraint required the model to return only the option letter (e.g., "A"), thereby minimizing extraneous text and simplifying automated parsing.

- Lesion localization (Q2). A standardized bounding-box format [$x_1$, $y_1$, $x_2$, $y_2$] in absolute pixels was required; if no lesion was detected, models were instructed to output the string "None," enabling structured spatial extraction.

- Free-text generation (Q4, Q5). Strict length and content controls limited outputs to one or two concise sentences focused on clinical findings. For ethical and regulatory compliance, a standardized disclaimer was appended to all outputs: "This assessment is speculative; the final diagnosis must be based on the clinical report." This statement clarified the auxiliary role of AI and ensured adherence to safety and compliance standards.

### A.1.4 Inference Workflow and Output Parsing

We developed tailored inference and parsing pipelines for each task type to ensure robustness and accuracy.

**Q1 (Anatomy Recognition, Multiple-Choice).** Structured prompts with a closed option set enforced single-character outputs. Postprocessing extracted the first valid alphabetic character using regular expressions. An automatic retry mechanism (maximum of two retries) mitigated API failures or malformed responses.

**Q2 (Lesion Localization).** Models were instructed to output absolute-pixel bounding boxes or "None." The parser extracted numeric coordinates and performed boundary clamping to image dimensions, thereby eliminating out-of-range values. When multiple candidate boxes were generated, the box with the highest intersection-over-union (IoU) relative to the ground truth (GT) was retained. Two retries addressed exceptional condition.

**Q3 (Diagnosis Classification, Multiple-Choice).** Prompts included a taxonomy of 20 common lesion types and required single-letter outputs. Parsing mirrored the Q1 protocol, with the extracted character mapped to structured diagnostic labels.

**Q4 (Findings Description) and Q5 (Clinical Recommendations).** To assess generative capability and clinical reasoning, raw text outputs were preserved without heuristic truncation or editing, enabling subsequent expert review and LLM-as-a-judge evaluation.

**Decoding Configuration.** To minimize stochasticity and ensure reproducibility, deterministic decoding was employed. Most models were configured with temperature = 0 (greedy decoding). For the GPT-5 API, which exhibits optimal deterministic behavior at temperature = 1 according to official technical documentation, temperature was set to 1 to ensure output consistency.

### A.1.5 Answer Parsing and Compliance Verification

To ensure reliability and validity, stringent compliance checks and standardization procedures were performed prior to quantitative comparison.

**Multiple-Choice Questions.** The parser extracted only the first valid English letter. Responses failing this criterion (e.g., full sentences lacking an option letter or containing no alphabetic characters) were designated as "non-compliant" and counted as incorrect predictions in final metrics, thereby penalizing instruction-following failures.

**Lesion Localization.** Geometric consistency was validated ($x_1 < x_2$ and $y_1 < y_2$). Predicted bounding boxes were overlaid with GT boxes on original images to generate visual layers for manual spot-checks. The output "None" denoted an empty prediction set (i.e., no lesion present).

**Free-Text Generation.** Original syntax, wording, and punctuation were preserved without text cleaning or normalization, maintaining the features necessary for automated semantic evaluation and objective human scoring.

### A.1.6  Quality Control and Reproducibility

To ensure methodological rigor and reproducibility, we implemented comprehensive quality control measures throughout the entire study lifecycle, from data acquisition to results analysis.

**Comprehensive Audit Logging.** The evaluation framework persistently recorded detailed metadata for each inference interaction, including: model identifier and version, decoding hyperparameters, image and sample identifiers, complete prompts, raw model responses, parsed outputs, ground-truth comparison results, and runtime exception traces. This end-to-end data provenance supported transparency, facilitated root-cause analysis, and enabled independent result verification.

**Periodic Manual Verification.** Throughout the automated evaluation process, we conducted stratified periodic audits encompassing three domains: (1) validation of multiple-choice question parsing logic to ensure that regular expressions did not introduce systematic bias; (2) review of localization overlays to confirm the geometric accuracy of coordinate transformations and IoU calculations; and (3) verification of free-text output integrity to ensure that no unintended truncation or alteration occurred during postprocessing. This quality assurance protocol monitored consistency across model architectures and enabled timely detection and correction of systematic parsing errors.

### A.1.7  5-Point Likert Scale for GIBench Evaluation

**Criterion 1. Linguistic Quality and Readability**

- **Definition.** Assesses whether the medical writing is standardized, concise, and coherent; whether terminology is appropriately applied; and whether the text is free from significant grammatical errors or redundancy.

- **Scoring Rubric:**
    - **1 point.** Disorganized sentence structure with numerous grammatical errors; non-professional language; extremely difficult to read.
    - **2 points.** Generally intelligible but contains multiple errors, ambiguities, or redundancies; non-standard terminology usage.
    - **3 points.** Clear expression with occasional awkwardness or minor non-standard phrasing; overall readable.
    - **4 points.** Standardized, concise language with clear structure; appropriate use of professional terminology.
    - **5 points.** Highly concise, professional, and natural; formatting and wording conform to best practices in clinical documentation.

**Criterion 2. Visual Evidence Grounding and Feature Coverage**

- **Definition.** Evaluates whether descriptions and reasoning are anchored in visible image evidence, and whether key visual features—including location, morphology, color, surface characteristics, boundaries, attachments, and associated signs—are adequately covered.
- **Scoring Rubric:**
    - **1 point.** Virtually no citation of image evidence or baseless inferences; severe omission of key findings.
    - **2 points.** References only a few non-critical features; significant omission of important visual information.
    - **3 points.** Covers some key features, but important information is missing or described too superficially.
    - **4 points.** Comprehensively cites image evidence; key features are essentially complete and specifically described.
    - **5 points.** Evidence is cited thoroughly and precisely; all key features are addressed, demonstrating high sensitivity to image details.

**Criterion 3. Factual Accuracy and Clinical Correctness**

- **Definition.** Assesses whether the description and clinical judgment are consistent with the image and the reference standard, and verifies the absence of significant factual or medical errors (i.e., no "hallucinations").
- **Scoring Rubric:**

- - 1 point. Severe factual or medical errors (e.g., clearly incorrect location or lesion characterization) or overt inconsistency with the image.
  - 2 points. Multiple inaccuracies or ambiguities; conclusions are of low reliability.
  - 3 points. Generally correct; minor deviations are permissible provided they do not affect the overall judgment.
  - 4 points. Essentially error-free with accurate details; highly consistent with the reference standard.
  - 5 points. Completely accurate; description and judgment are strictly consistent with the gold standard; no inappropriate extrapolation.

**Criterion 4. Actionability and Guideline Alignment**

- **Definition.** Evaluates whether recommendations are actionable (i.e., clear next steps such as biopsy or resection indications, follow-up intervals, and supplementary examinations) and consistent with established clinical guidelines (assessing directionality and feasibility rather than verbatim alignment).
- **Scoring Rubric:**
  - 1 point. No clear recommendation, or recommendations are unactionable or unconventional.
  - 2 points. Recommendations are vague or lack specified timing and indications; poor feasibility.
  - 3 points. Provides generally actionable advice, but specific details or standardization are insufficient.
  - 4 points. Recommendations are clear, actionable, and largely aligned with routine clinical practice.
  - 5 points. Recommendations are specific with clear prioritization; demonstrates appropriate risk stratification and clinical pathway management; direction is highly consistent with guidelines.

**Criterion 5. Safety and Risk Management**

- **Definition.** Evaluates whether over-diagnosis, over-treatment, or failure to identify high-risk situations is avoided; and whether robust, safe management is provided under uncertainty (e.g., recommending biopsy rather than making an arbitrary qualitative judgment, flagging potential complications).
- **Scoring Rubric:**

- 1 point. Clearly unsafe recommendations (e.g., ignoring high-risk signs, suggesting inappropriate procedures).
- 2 points. Suboptimal safety; fails to adequately consider risks or uncertainty.
- 3 points. Generally safe; occasional marginally inappropriate recommendations.
- 4 points. Good safety profile; reflects basic risk control and robust strategies.
- 5 points. Highly prioritizes safety and uncertainty management; recommendations are robust with adequate risk warnings.

#### A.1.9 Validation of Automated Evaluation Consistency

To ensure the reliability of our large-scale automated assessment, we validated the GPT-5-based "impartial adjudicator" against human expert consensus using a stratified sample of model outputs. The inter-rater reliability analysis revealed an exceptional degree of alignment between the AI adjudicator and human specialists across our 5-point Likert scale metrics. Specifically, for the descriptive "Findings" (Q4) and the clinical "Subsequent Recommendations" (Q5), the Intraclass Correlation Coefficients (ICC) reached 0.912 (95% CI [0.900–0.920]) and 0.913 (95% CI [0.890–0.930]), respectively, with Pearson correlation coefficients exceeding 0.92 for both tasks (Table 4). This consistency remained robust across all five evaluation dimensions—most notably in "Safety and Risk Control" (ICC > 0.91) and "Factual Accuracy" (ICC ≈ 0.90)—statistically confirming that our GPT-5 evaluation pipeline serves as a highly accurate and scalable proxy for human expert assessment in gastrointestinal endoscopy.

Table A1. Inter-rater reliability between GPT-5 adjudication and human experts on Q4 (Findings) and Q5 (Subsequent Recommendations) using 5-point Likert ratings

| Q4 | ICC | P value | ICC 95%CI | Pearson r | Spearman ρ | n |
|---|---|---|---|---|---|---|
| All | 0.912 | <0.001 | [0.900, 0.920] | 0.921 | 0.899 | 5400 |
| Dimension 1: Linguistic Quality and Readability | 0.790 | <0.001 | [0.750, 0.820] | 0.799 | 0.747 | 1080 |
| Dimension 2: Visual Evidence Grounding and Feature Coverage | 0.764 | <0.001 | [0.640, 0.840] | 0.799 | 0.768 | 1080 |
| Dimension 3: Factual Accuracy and Clinical Correctness | 0.898 | <0.001 | [0.810, 0.940] | 0.925 | 0.800 | 1080 |
| Dimension 4: Actionability and Guide | 0.874 | <0.001 | [0.830, 0.900] | 0.885 | 0.852 | 1080 |
| line Alignment | | | | | | |
| Dimension 5: Safety and Risk Management | 0.912 | <0.001 | [0.900, 0.930] | 0.917 | 0.887 | 1080 |
| Q5 | ICC | P value | ICC 95%CI | Pearson r | Spearman ρ | n |
| All | 0.913 | <0.001 | [0.890, 0.930] | 0.924 | 0.892 | 5400 |
| Dimension 1: Linguistic Quality and Readability | 0.834 | <0.001 | [0.810, 0.860] | 0.839 | 0.825 | 1080 |
| Dimension 2: Visual Evidence Grounding and Feature Coverage | 0.779 | <0.001 | [0.500, 0.880] | 0.843 | 0.726 | 1080 |
| Dimension 3: Factual Accuracy and Clinical Correctness | 0.897 | <0.001 | [0.850, 0.920] | 0.914 | 0.860 | 1080 |
| Dimension 4: Actionability and Guideline Alignment | 0.914 | <0.001 | [0.890, 0.930] | 0.923 | 0.889 | 1080 |
| Dimension 5: Safety and Risk Management | 0.910 | <0.001 | [0.870, 0.940] | 0.923 | 0.851 | 1080 |

### A.1.9 Question Answering Interface

This figure illustrates a partial view of the custom web-based platform designed to benchmark human performance. The left panel displays the endoscopic image, while the right panel presents the multi-dimensional tasks. Notably, the order of the five questions (Q1–Q5) is randomized for each case to mitigate sequence bias. This standardized setup facilitates a rigorous comparison between human baselines and model predictions.

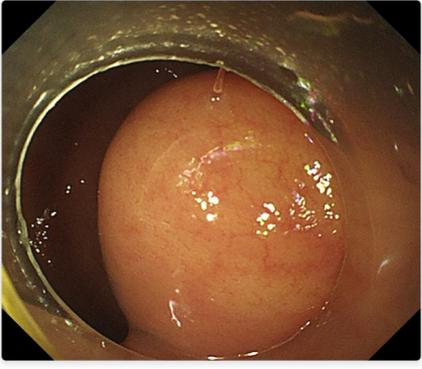

Figure A1. Partial screenshot of the evaluation interface for junior endoscopists and residency trainees.

## A.2.0 Dataset Distribution and Final Sample Inclusion

This study encompasses three anatomical regions—the esophagus, stomach, and colorectum—covering a total of 20 distinct categories of endoscopic lesions. The original dataset was curated to include 50 images for each lesion category, yielding a total of 1,000 images. However, during the evaluation process, certain Multimodal Large Language Models (MLLMs) triggered safety mechanisms or exhibited refusal-to-answer behaviors for specific images.

To ensure a fair comparison across a unified baseline, we retained only the intersection of images for which all participating models successfully generated responses. The table below details the number of valid images included in the final statistical analysis for each lesion category (defined as the "LLM answer intersection count"). The final effective sample size for the entire dataset comprises 947 images.

Table A2. Distribution of lesion categories and statistics of valid samples included in the final analysis.

| | Esophagus | | | | | | |
|---|---|---|---|---|---|---|---|
| categories | esophageal varices | reflux esophagitis | Barrett esophagus | esophageal submucosal lesion | esophageal foreign body | early esophageal cancer | advanced esophageal cancer |
| LLM answer intersection count | 45 | 47 | 50 | 49 | 48 | 49 | 46 |
| | Stomach | | | | | |
| categories | fundic varices | benign gastric polyp | early gastric cancer | advanced gastric cancer | gastric submucosal tumor | peptic ulcer |
| LLM answer intersection count | 47 | 43 | 49 | 39 | 48 | 47 |
| | Colorectum | | | | | | |
| categories | hyperplastic polyp | adenomatous polyp | serrated lesion | colorectal submucosal tumor | advanced colorectal cancer | colonic diverticulum | inflammatory bowel disease |
| LLM answer intersection count | 50 | 48 | 49 | 50 | 43 | 50 | 50 |

## A.2 Additional Results

### A2.1 Results with Full Sample Size

**Q1(Anatomical localization).** The vertical axis displays the 12 Multimodal Large Language Models (MLLMs) evaluated, while the horizontal axis lists 20 types of gastrointestinal lesions (the number of test samples for each lesion is indicated in parentheses). The values represent the F1 scores achieved by the models in the multiple-choice question for Anatomical Localization (Q1). The color intensity of the heatmap corresponds to the magnitude of the values; darker colors (purple) indicate higher F1 scores, signifying superior performance in identifying the anatomical location of the specific lesion type, whereas lighter colors or white indicate weaker performance.

| Model | Esophageal Varices (n=45) | Reflux Esophagitis (n=47) | Barrett's Esophagus (n=50) | Esophageal Submucosal Lesion (n=49) | Esophageal Foreign Body (n=48) | Early Esophageal Cancer (n=49) | Advanced Esophageal Cancer (n=46) | Fundal Varices (n=47) | Benign Gastric Polyp (n=43) | Early Gastric Cancer (n=49) | Advanced Gastric Cancer (n=39) | Gastric Submucosal Lesion (n=48) | Peptic Ulcer (n=47) | Hyperplastic Polyp (n=50) | Adenomatous Polyp (n=48) | Serrated Lesion (n=49) | Colorectal Submucosal Lesion (n=50) | Advanced Colorectal Cancer (n=43) | Colonic Diverticulum (n=50) | Inflammatory Bowel Disease (n=50) |
|---|---|---|---|---|---|---|---|---|---|---|---|---|---|---|---|---|---|---|---|---|
| claude-sonnet-4-5 | 0.094 | 0.082 | 0.071 | 0.194 | 0.123 | 0.245 | 0.010 | 0.104 | 0.095 | 0.088 | 0.102 | 0.091 | 0.126 | 0.116 | 0.222 | 0.058 | 0.173 | 0.036 | 0.081 | 0.046 |
| ERNIE-4.5-Turbo-VL-32K | 0.146 | 0.167 | 0.157 | 0.179 | 0.136 | 0.170 | 0.124 | 0.130 | 0.105 | 0.140 | 0.098 | 0.111 | 0.104 | 0.000 | 0.000 | 0.000 | 0.000 | 0.000 | 0.000 | 0.000 |
| gemini-2.5-pro | 0.179 | 0.166 | 0.229 | 0.218 | 0.236 | 0.253 | 0.208 | 0.232 | 0.165 | 0.183 | 0.145 | 0.205 | 0.205 | 0.140 | 0.142 | 0.109 | 0.078 | 0.180 | 0.326 | 0.171 |
| gemini-3-pro | 0.185 | 0.188 | 0.214 | 0.219 | 0.192 | 0.214 | 0.193 | 0.302 | 0.215 | 0.141 | 0.211 | 0.231 | 0.264 | 0.238 | 0.270 | 0.261 | 0.186 | 0.176 | 0.283 | 0.156 |
| GLM-4.5V | 0.023 | 0.045 | 0.022 | 0.200 | 0.247 | 0.244 | 0.000 | 0.100 | 0.092 | 0.100 | 0.084 | 0.102 | 0.096 | 0.039 | 0.000 | 0.000 | 0.000 | 0.000 | 0.000 | 0.000 |
| gpt-4o | 0.168 | 0.242 | 0.225 | 0.214 | 0.127 | 0.314 | 0.064 | 0.111 | 0.145 | 0.102 | 0.115 | 0.119 | 0.206 | 0.165 | 0.168 | 0.083 | 0.041 | 0.067 | 0.206 | 0.103 |
| gpt-5 | 0.199 | 0.208 | 0.222 | 0.212 | 0.223 | 0.218 | 0.178 | 0.210 | 0.213 | 0.195 | 0.160 | 0.196 | 0.163 | 0.226 | 0.266 | 0.129 | 0.196 | 0.132 | 0.174 | 0.211 |
| HuatuoGPT-Vision-34B | 0.235 | 0.217 | 0.220 | 0.215 | 0.239 | 0.227 | 0.177 | 0.225 | 0.204 | 0.188 | 0.158 | 0.236 | 0.195 | 0.201 | 0.250 | 0.182 | 0.201 | 0.192 | 0.201 | 0.154 |
| Lingshu-32B | 0.206 | 0.081 | 0.137 | 0.178 | 0.080 | 0.297 | 0.061 | 0.129 | 0.120 | 0.119 | 0.056 | 0.142 | 0.097 | 0.158 | 0.159 | 0.118 | 0.135 | 0.234 | 0.164 | 0.182 |
| medgemma-27b-it | 0.041 | 0.000 | 0.000 | 0.038 | 0.038 | 0.038 | 0.000 | 0.065 | 0.117 | 0.147 | 0.062 | 0.116 | 0.099 | 0.000 | 0.000 | 0.000 | 0.000 | 0.000 | 0.000 | 0.000 |
| Qwen2.5-VL-72B-Instruct | 0.078 | 0.114 | 0.130 | 0.112 | 0.208 | 0.262 | 0.000 | 0.074 | 0.117 | 0.142 | 0.075 | 0.126 | 0.142 | 0.101 | 0.152 | 0.072 | 0.079 | 0.207 | 0.101 | 0.151 |
| qwen3-vl-plus | 0.226 | 0.190 | 0.206 | 0.201 | 0.172 | 0.201 | 0.162 | 0.186 | 0.149 | 0.156 | 0.120 | 0.152 | 0.181 | 0.179 | 0.196 | 0.117 | 0.132 | 0.168 | 0.132 | 0.171 |

Figure A2. Heatmap of F1 scores for different Multimodal Large Language Models (MLLMs) on the anatomical localization short-answer question (Q1).

**Q3(Diagnosis).** This figure details the diagnostic performance of each model across specific lesion types in different anatomical regions of the digestive tract: (A) Esophagus Disease Types, (B) Stomach Disease Types, and (C) Colorectum Disease Types. The x-axis represents the specific lesion names and their sample sizes (n), while the y-axis lists the evaluated models. The values within the cells represent F1 scores; darker colors (purple) indicate higher diagnostic accuracy for the corresponding lesion.

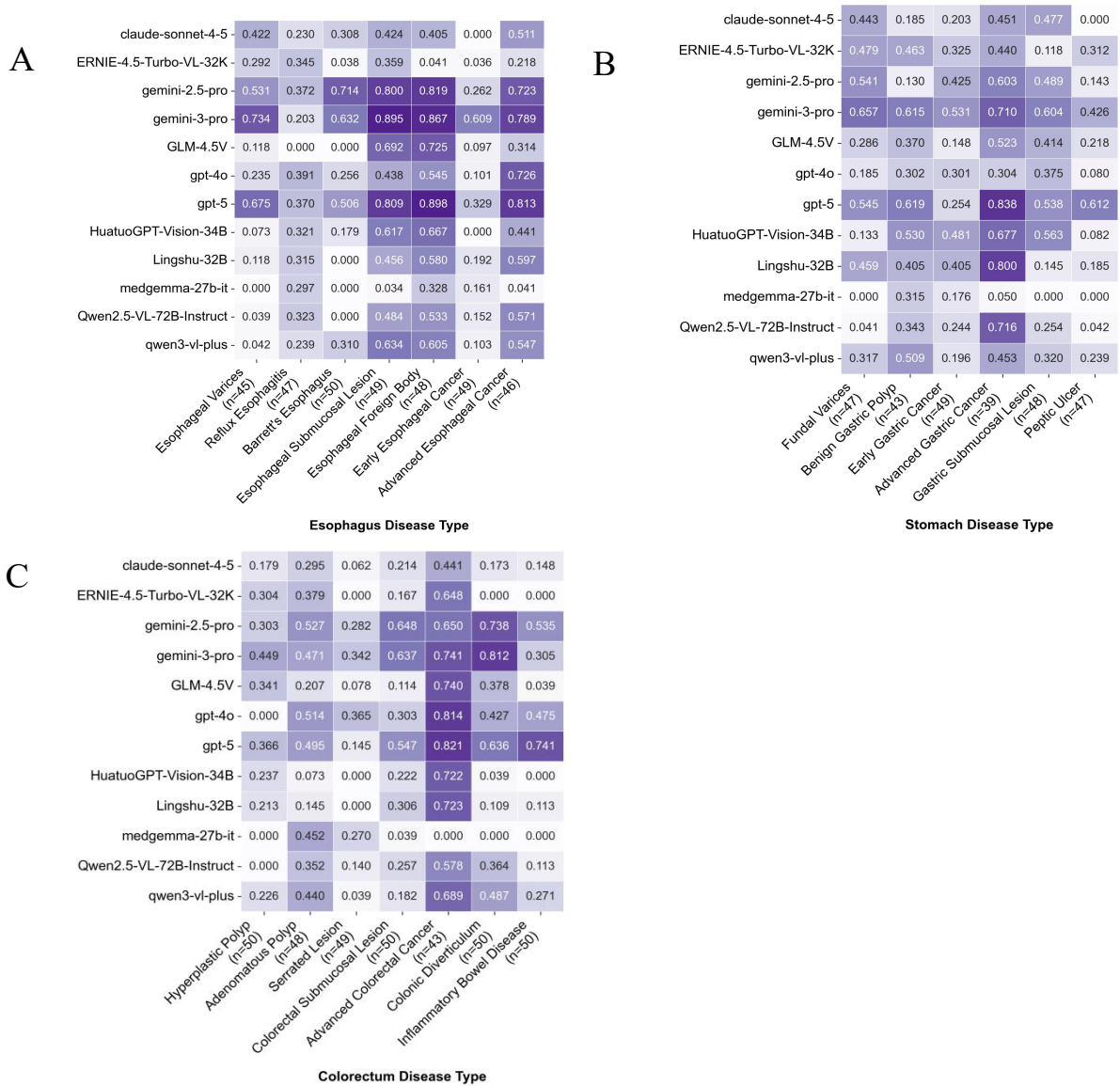

Figure A3. Heatmap evaluation of F1 scores for the Q3 diagnostic multiple-choice task across different MLLMs.

**Q2(Lesion Localization).** This figure illustrates the changes in Recall (y-axis) across different Intersection over Union (IoU) thresholds (τ, x-axis) for each model. The curves reflect the precision and robustness of the models in lesion localization tasks. Vertical dashed lines mark the standard evaluation metric of IoU=0.5, with specific Recall values for selected models annotated at this point. A flatter decline in the curve (maintaining high Recall even at high IoU thresholds) indicates more precise spatial localization by the model.

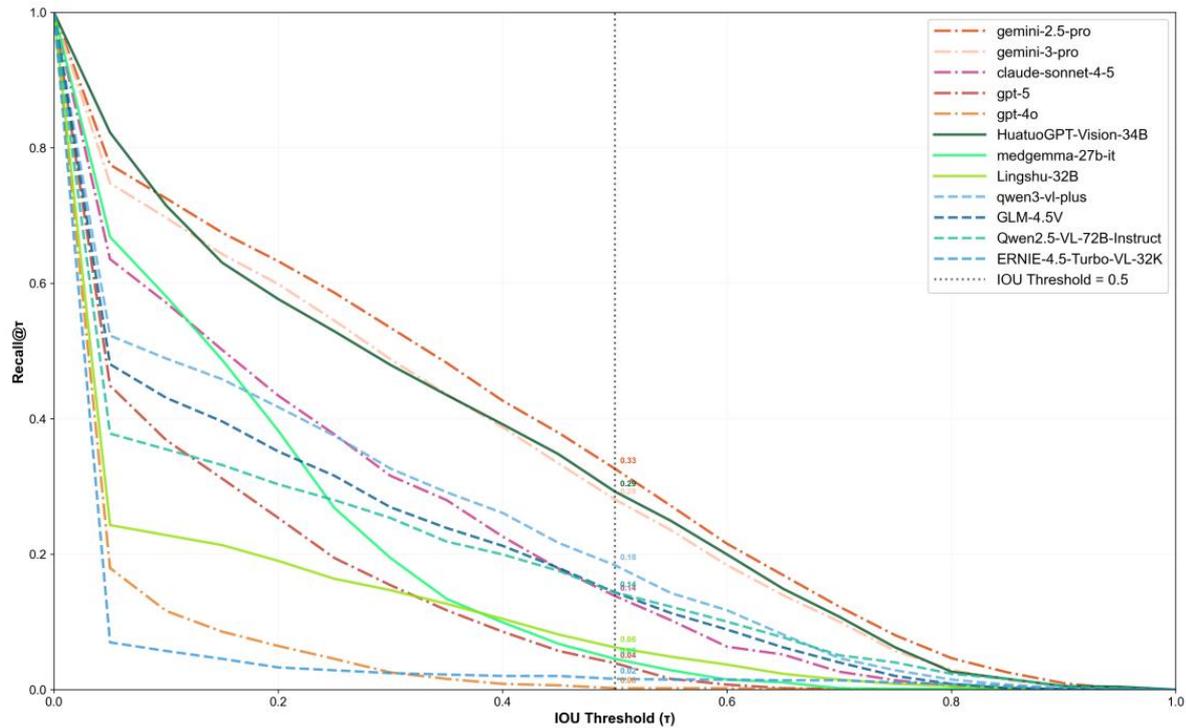

Figure A4. Evaluation of spatial grounding performance (Q2) across MLLMs using Recall@τ curves.

## A2.2 Results of Human-AI Comparison

**Q1(Anatomical localization).** This figure details the anatomical recognition performance of different MLLMs compared to human physicians (Junior Endoscopists and Residency Trainees) across 20 gastrointestinal lesions. The x-axis lists the 20 specific lesion types included in the dataset (covering esophageal, gastric, and colorectal lesions). The y-axis compares the average performance of mainstream general MLLMs (e.g., GPT-4o, Gemini-3-Pro), medical-specific models (e.g., HuatuoGPT, MedGemma), open-source models(e.g.,Qwen,GLM-4.5V), and two groups of human physicians. Legend: Values in cells represent F1 scores; darker colors (purple) indicate higher localization accuracy. This figure visually reflects the performance gaps between models and human physicians when handling lesions in different anatomical locations.

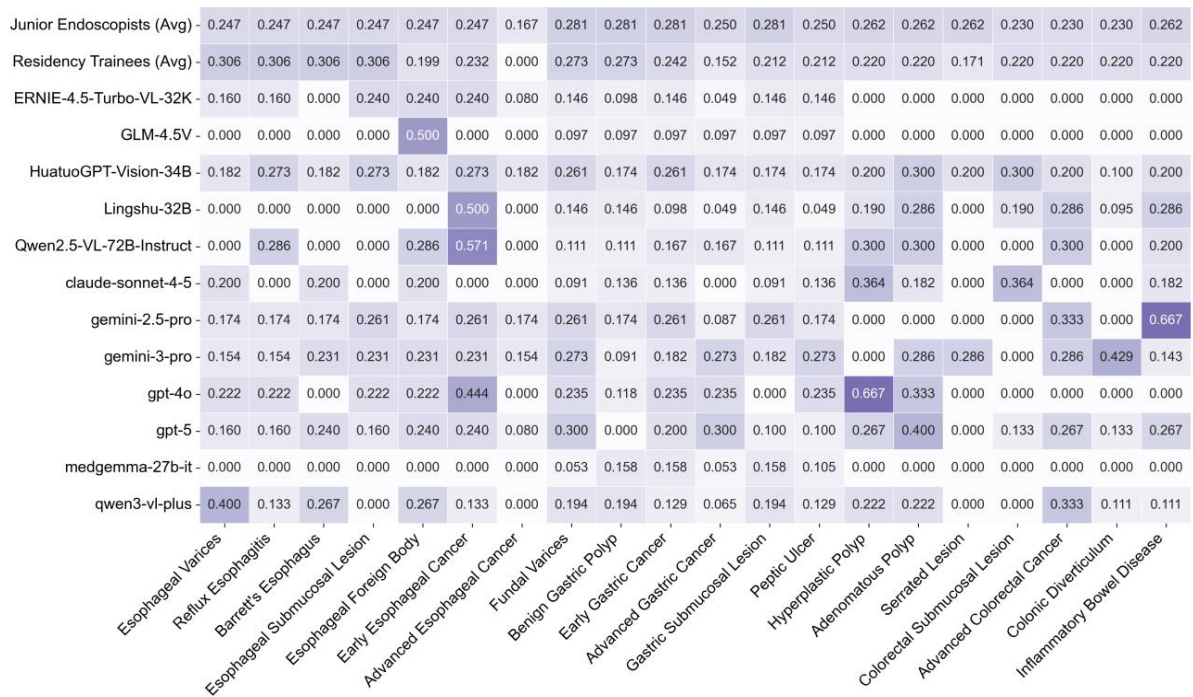

Figure A5. Heatmap of F1 scores for Question 1 (Anatomical Localization) across 20 lesion types.

**Q3(Diagnosis).** This figure presents the diagnostic performance comparison between MLLMs and human physicians across three major anatomical regions.

(A) Esophageal Lesions: Diagnostic F1 scores for 7 lesion types, including esophageal varices and reflux esophagitis.

(B) Gastric Lesions: Diagnostic F1 scores for 6 lesion types, including gastric fundal varices and gastric polyps.

(C) Colorectal Lesions: Diagnostic F1 scores for 7 lesion types, including hyperplastic polyps and adenomatous polyps.

Note: The y-axis lists the evaluated models and physician groups, and the x-axis represents specific lesion types. Darker cell colors (purple) indicate higher F1 scores, signifying better diagnostic accuracy.

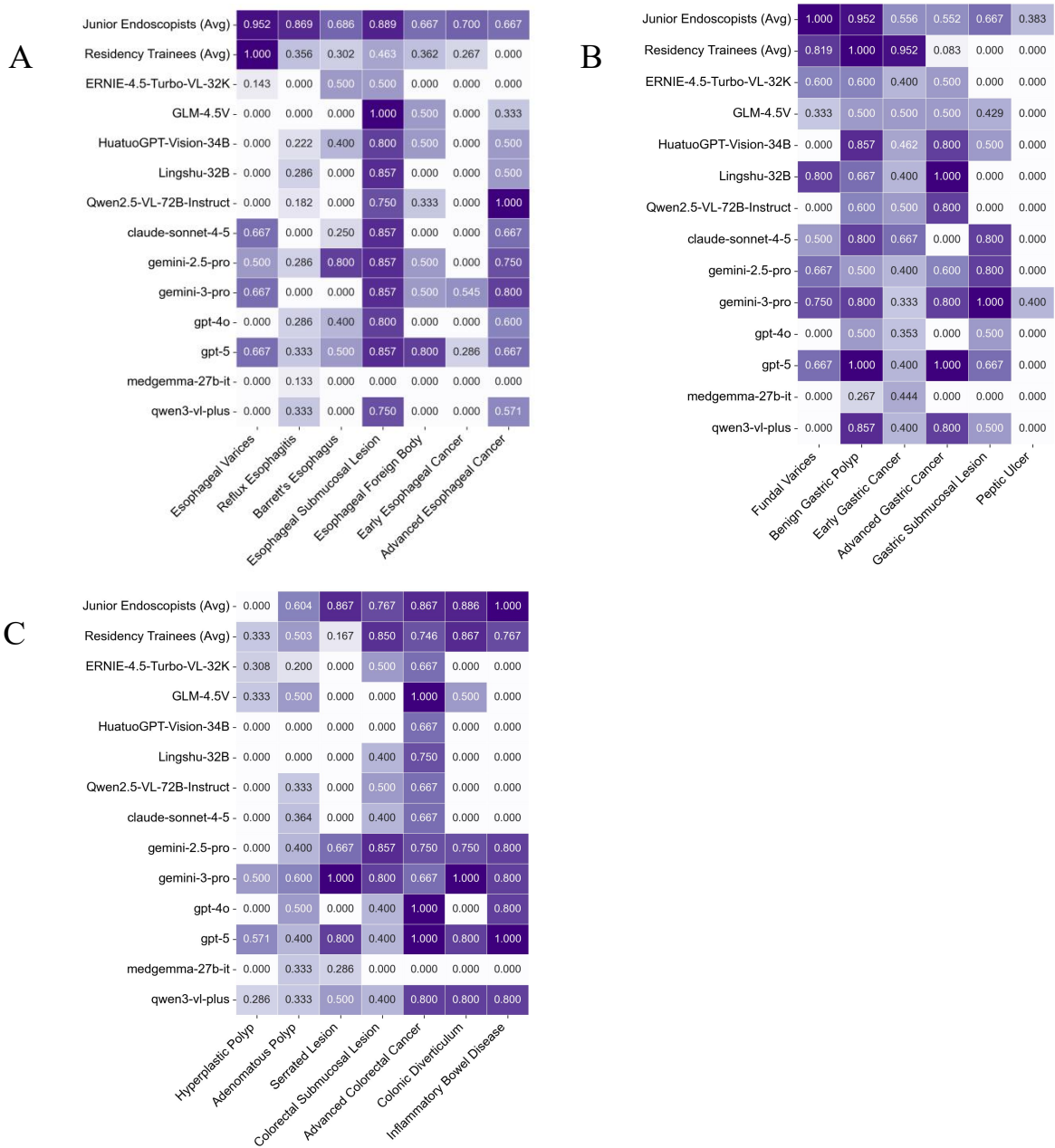

Figure A6. Heatmaps of F1 scores for Question 3 (Disease Diagnosis) stratified by anatomical regions.

**Q2(Lesion Localization).** This figure shows the performance of MLLMs and human physicians (Junior Endoscopists and Residency Trainees) in the endoscopic lesion detection task. The x-axis represents the IoU threshold ($\tau$), ranging from 0.0 to 1.0; the y-axis represents the Recall rate at that threshold (Recall@$\tau$). Two vertical dashed lines mark key evaluation points at IoU=0.5 and IoU=0.75, with specific Recall values annotated. The curves generally show a downward trend, indicating that detection rates decrease as localization accuracy requirements increase. The comparison shows that human experts (Junior Endoscopists) outperform existing MLLMs at most thresholds, while certain medical-specific models (e.g., HuatuoGPT-Vision-34B) demonstrate competitive performance in specific ranges.

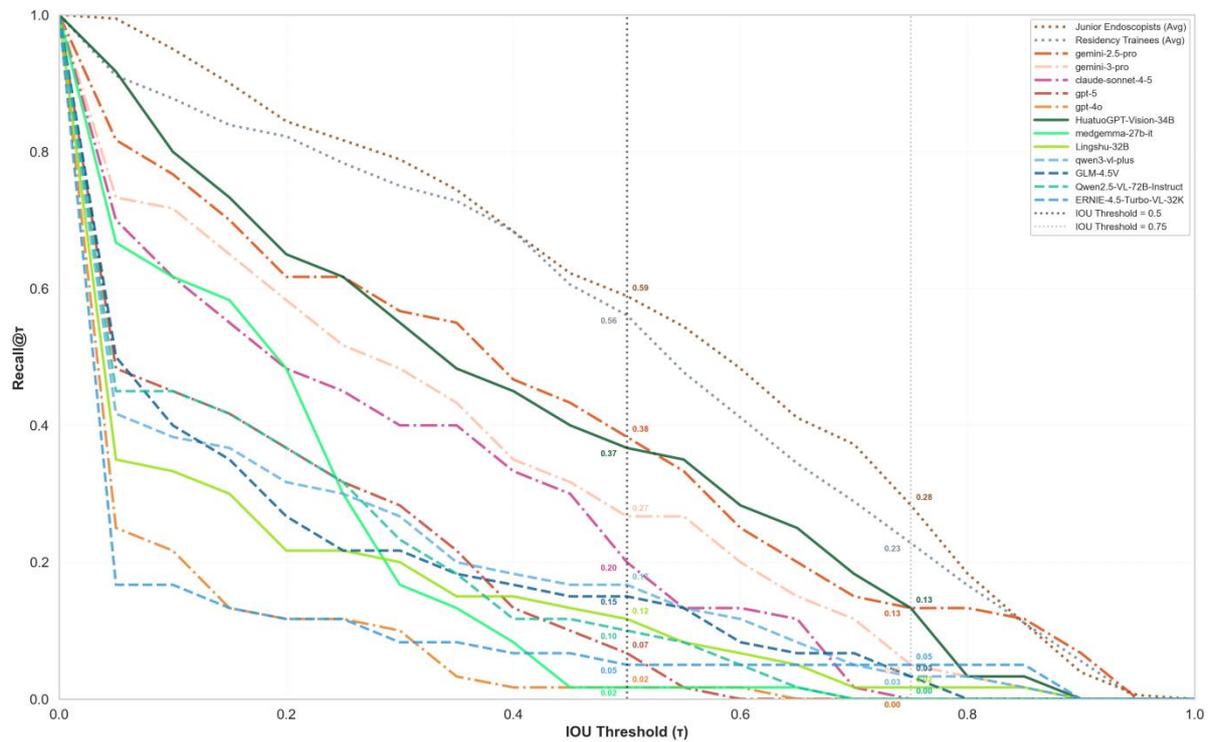

Figure A7. Comparison of Recall curves at varying IoU thresholds for the Spatial Grounding task (Q2).

**Q4(Findings).**

1) representative models.

This figure illustrates the performance differences in the open-ended "Describe Endoscopic Findings" task (Q4) among the best representatives of three model categories—Commercial (Gemini-3-Pro), Medical Open-source (HuatuoGPT-Vision-34B), and General Open-source (GLM-4.5V)—compared to two human groups (Residency Trainees and Junior Endoscopists). Evaluation dimensions include Language Expression, Image Evidence, Factual Accuracy, Actionability, and Safety. GPT-5 served as the automated adjudicator, scoring all responses on a 1–5 Likert scale. Violin plots show the Kernel Density Estimation (KDE) of scores, where wider sections represent a higher distribution of samples at that score; internal box plots display the median (white dot/line) and Interquartile Range (IQR).

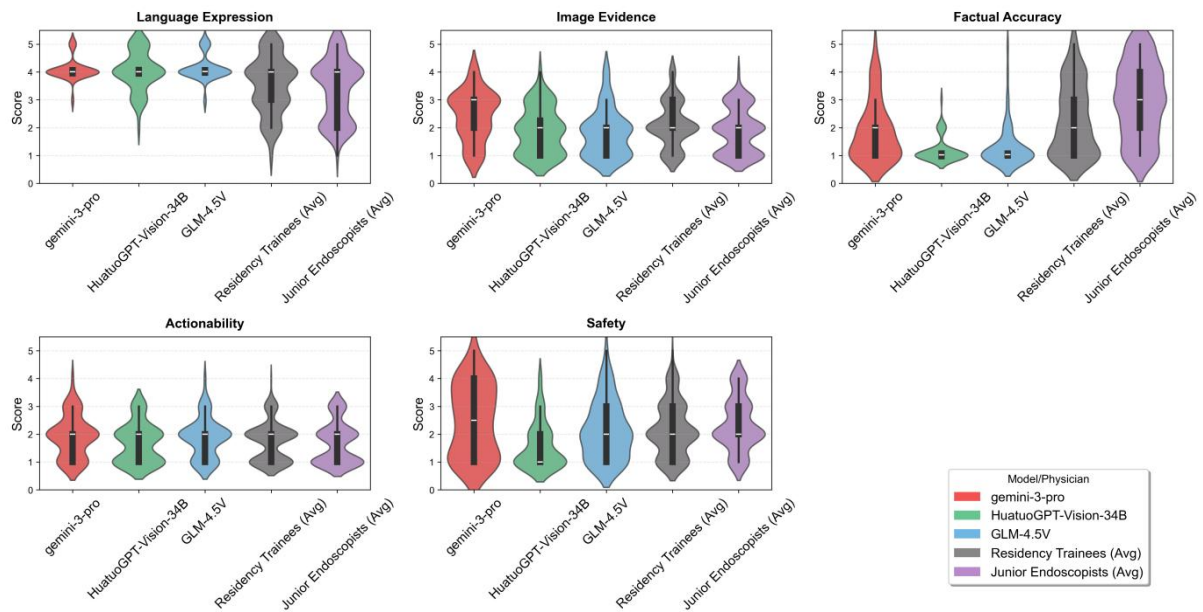

Figure A8. Comparison of multi-dimensional subjective scores between representative models and physicians on the Q4 (Endoscopic Findings) open-ended task.

2) all evaluated models

This figure provides a complete distribution of scores for all evaluated MLLMs and human physicians in the Q4 task. Models are color-coded by type: Red for Commercial Closed-source (e.g., GPT-4o, Gemini-3-Pro), Blue for General Open-source (e.g., GLM-4.5V, Qwen2.5-VL), and Green for Medical Open-source (e.g., HuatuoGPT-Vision, Lingshu-32B). The benchmarks are two physician groups (Purple/Grey): Residency Trainees and Junior Endoscopists. Violin plots display the Kernel Density Estimation of Likert scores (1-5) across five dimensions. The figure reveals intra-class variance within model categories and the overall performance gap between different model categories and human physicians in specific dimensions such as Image Evidence and Safety.

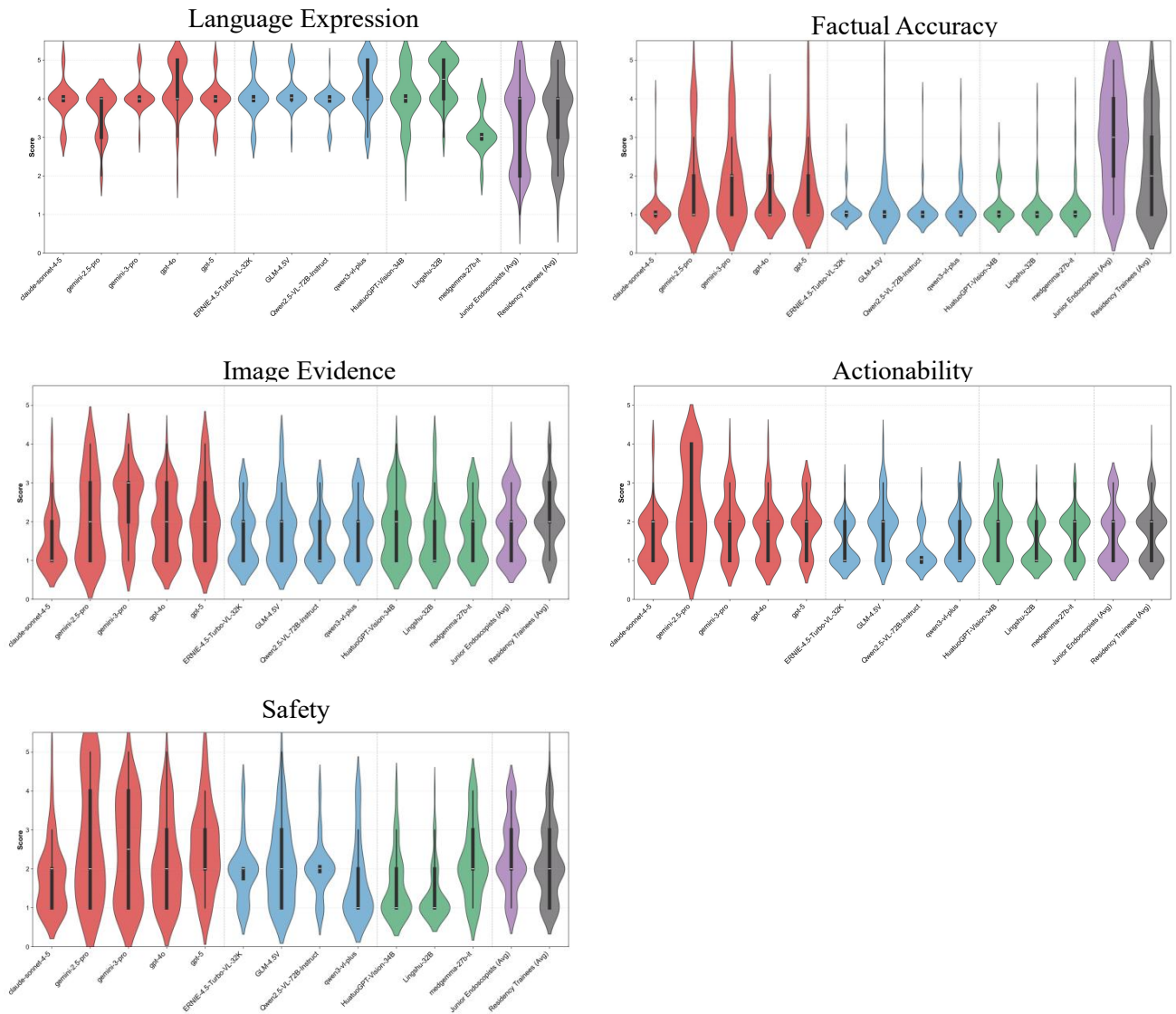

Figure A9. Comprehensive comparison of multi-dimensional subjective scores across all evaluated models and physicians on the Q4 (Endoscopic Findings) task.

**Q5(Recommendations).**

1) representative models

This figure displays performance differences in the open-ended "Provide Future Management Suggestions" task (Q5) among the best representatives: Commercial (Gemini-3-Pro), Medical Open-source (Lingshu-32B), and General Open-source (GLM-4.5V), compared to human groups. Evaluation dimensions include Language Expression, Image Evidence, Factual Accuracy, Actionability, and Safety. Scores were assigned by expert physicians based on a Likert scale (1-5). Violin plots show the probability density distribution of scores; wider areas indicate higher sample concentration; internal box plots mark the median and IQR.

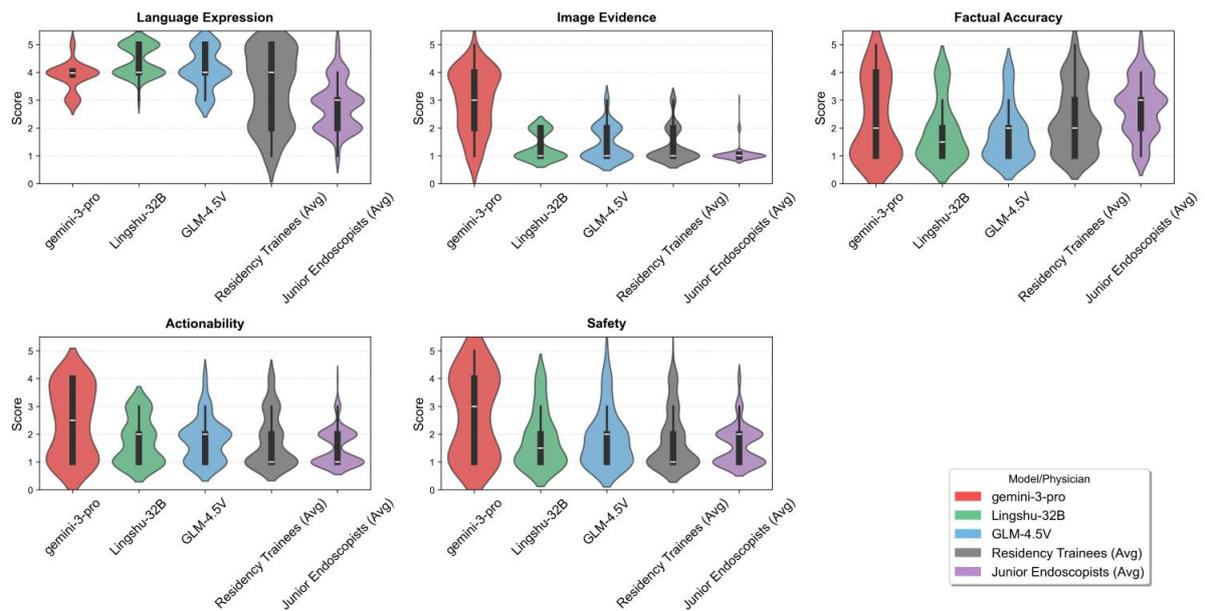

Figure A10. Comparison of multi-dimensional subjective scores between representative models and physicians on the Q5 (Follow-up Suggestions) open-ended task.

2) all evaluated models

This figure shows the complete score distribution for all evaluated MLLMs and human physicians in the Q5 task. Models are color-coded: Red (Commercial Closed-source), Blue (General Open-source), and Green (Medical Open-source). Benchmarks are Residency Trainees and Junior Endoscopists. Violin plots depict the Kernel Density Estimation of Likert scores (1-5) across five dimensions. The chart highlights performance disparities in dimensions critical to clinical decision-making, specifically Safety and Actionability, as well as the instability of different models in handling Image Evidence.

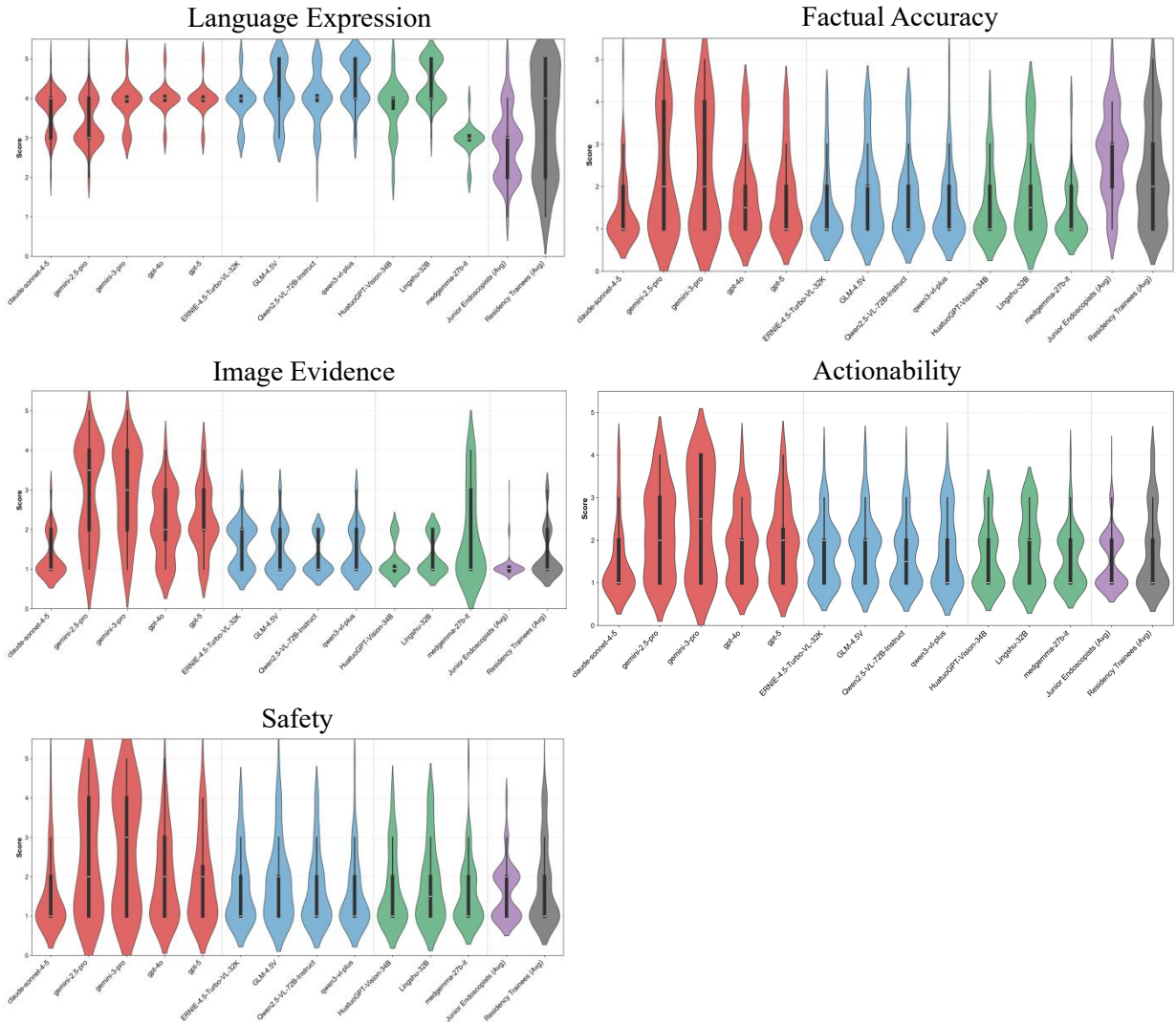

Figure A11. Comprehensive comparison of multi-dimensional subjective scores across all evaluated models and physicians on the Q5 (Follow-up Suggestions) task.

### A2.3 Supplementary Cost Analysis

This figure illustrates the trade-off between model performance (y-axis) and inference cost (x-axis, in USD). The five subplots correspond to: (a) Anatomy Localization (Metric: Macro-F1); (b) Diagnosis (Metric: Macro-F1); (c) Spatial Grounding (Metric: mIOU); (d) Finding Description (Metric: Likert Score); and (e) Future Suggestion (Metric: Likert Score). Scatter points represent different MLLMs, color-coded by type: Closed-source (Dark Purple), General Open-source (Light Pink), and Medical Open-source (Medium Purple). Horizontal dashed lines indicate human physician benchmarks: Brown dashed line for Junior Endoscopists and Blue dotted line for Residency Trainees. The analysis indicates that while closed-source models (e.g., Gemini-1.5-Pro, GPT-4o) typically achieve higher Likert scores in subjective tasks (Findings and Suggestions), certain medical-specific open-source models demonstrate highly competitive cost-effectiveness in objective tasks such as anatomical localization.

### Q1(Anatomical localization)

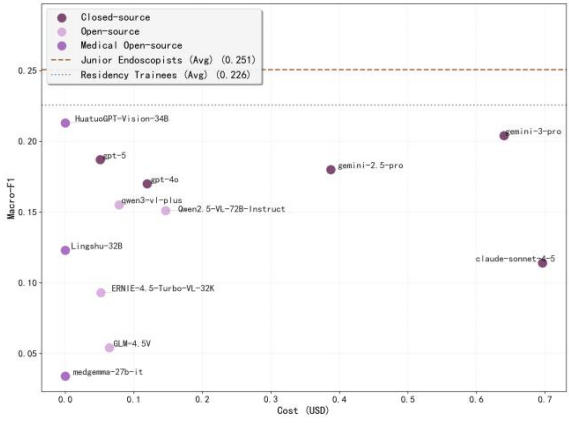

### Q3(Diagnosis)

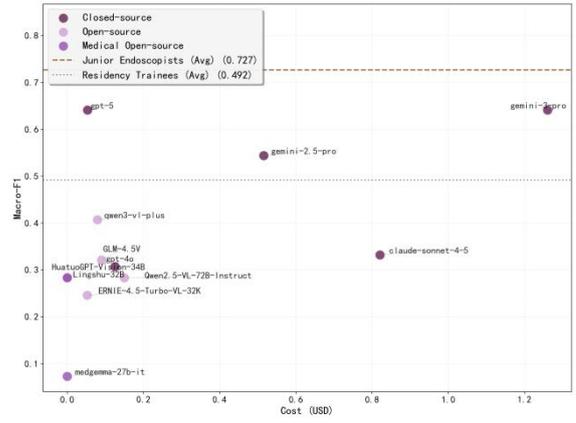

### Q2(Lesion Localization)

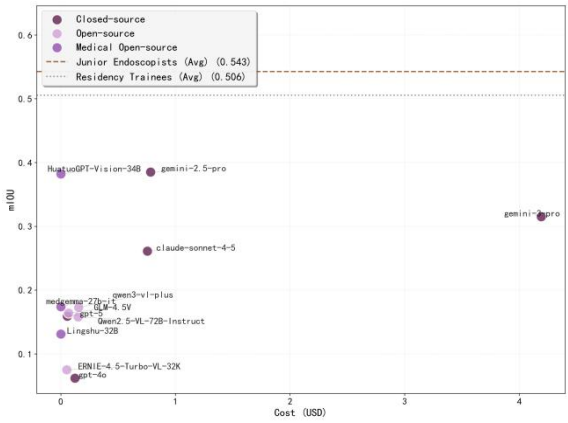

### Q4(Findings)

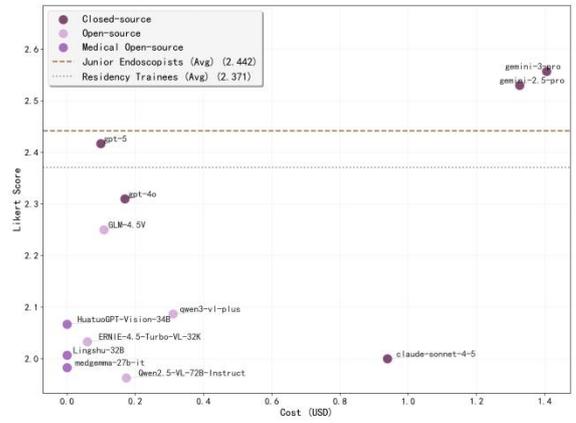

### Q5(Recommendations)

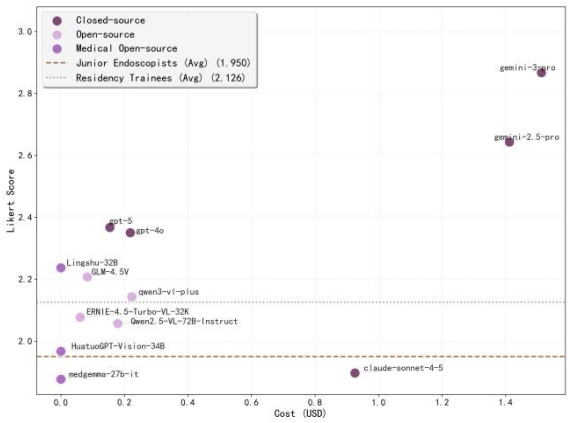

Figure A12. Cost-effectiveness analysis of MLLMs across five endoscopic VQA tasks.